\documentclass[journal ]{new-aiaa}
\usepackage[utf8]{inputenc}
\usepackage{textcomp}

\usepackage{appendix}
\usepackage{subfig}

\usepackage{graphicx}
\usepackage[dvipsnames]{xcolor}
\usepackage{amsmath}
\usepackage[version=4]{mhchem}
\usepackage{siunitx}
\usepackage{longtable,tabularx}
\usepackage{multicol}
\usepackage{gensymb}
\usepackage{soul}
\usepackage{orcidlink}
\newcommand{\myequations}[1]{%
\addcontentsline{equ}{myequations}{\protect\numberline{\theequation}#1}\par}

\title{Explainable Artificial Intelligence for Exhaust Gas Temperature of Turbofan Engines}

\author{Marios Kefalas \footnote{PhD Candidate, LIACS, email: m.kefalas@liacs.leidenuniv.nl. (Corresponding Author).}~\orcidlink{0000-0002-2422-758X} and Juan de Santiago Rojo Jr.\footnote{MSc student, juan.desantiagorojo@gmail.com.}}
\affil{LIACS, Leiden University, 2333 CA, Leiden, South Holland, The Netherlands}
\author{Asteris Apostolidis\footnote{Technology Innovation Manager, Strategy Technology Office, KLM, asterisa@gmail.com, AIAA Professional Member.}~\orcidlink{0000-0002-4083-8348} and Dirk van den Herik \footnote{Lead Engineer on engines and engine performance, On-wing Engineering, KLM, dirk-van-den.herik@klm.com.}}
\affil{KLM Royal Dutch Airlines, P.O. Box 7700, 1117 ZL, Schiphol, North Holland, The Netherlands}
\author{Bas van Stein\footnote{Researcher, LIACS, email: b.van.stein@liacs.leidenuniv.nl.}~\orcidlink{0000-0002-0013-7969} and Thomas B\"ack\footnote{Professor of Natural Computing, LIACS, email: t.h.w.baeck@liacs.leidenuniv.nl.}~\orcidlink{0000-0001-6768-1478}}
\affil{LIACS, Leiden University, 2333 CA, Leiden, South Hollad, The Netherlands}

\begin{document}

\maketitle

\begin{abstract}

Data-driven modeling is an imperative tool in various industrial applications, including many applications in the sectors of aeronautics and commercial aviation.
These models are in charge of providing key insights, such as which parameters are important on a specific measured outcome or which parameter values we should expect to observe given a set of input parameters. 
At the same time, however, these models rely heavily on assumptions (e.g., stationarity) or are ``black box'' (e.g., deep neural networks), meaning that they lack interpretability of their internal working and can be viewed only in terms of their inputs and outputs.
An interpretable alternative to the ``black box" models and with considerably less assumptions is symbolic regression (SR).
SR searches for the optimal model structure while simultaneously optimizing the model's parameters without relying on an a-priori model structure.
In this work, we apply SR on real-life exhaust gas temperature (EGT) data, collected at high frequencies through the entire flight, in order to uncover meaningful algebraic relationships between the EGT and other measurable engine parameters.
The experimental results exhibit promising model accuracy, as well as explainability returning an absolute difference of $3$\textdegree C compared to the ground truth and demonstrating consistency from an engineering perspective.

\end{abstract}

\section{Introduction}\label{Introduction}

\lettrine{D}{ata}-driven modeling is an imperative tool in various industrial applications, including many applications in the sectors of aeronautics and commercial aviation. 
By data-driven modeling, we do not imply a conceptual model that is based on the data requirements of an application being developed, but rather a model of the underlying data-generating process. 
Such predictive models perform the task of identifying complex patterns in 
multimodal data, something that can also be loosely termed as reverse engineering~\cite{vaddireddy_feature_2020}.
These models are in charge of providing key insights, such as which parameters (covariates) are important on a specific measured outcome or which parameter values we should expect to observe given a set of input parameters. 
In addition, such models can infer future states of the system and distill new or refine existing physical models of nonlinear dynamical systems~\cite{vaddireddy_feature_2020}.

A physics-based model that adequately fits the data requires
a thorough understanding of the system’s physics and processes, which
can be prohibitively costly in terms of time and resources. 
On the other hand, there are cases where such models are necessary, especially in applications where no sufficient data have been generated yet. 
A good example is the design and certification phase of new aeronautical systems. 
Linear and nonlinear statistical models rely on assumptions that might not hold (e.g., stationarity for ARMA ~\cite{box2015time} models in case of time-series). 
In contrast to those, non-parametric machine learning (ML) algorithms, such as the more recent deep neural networks (DNNs)~\cite{ian_goodfellow_deep_2016}
are considered to be ``black box'' models,
referring to processes  which  lack interpretability of their internal workings and can be viewed only in terms of their inputs and outputs. 
This means that these models do not explain their predictions/outputs in a way that is understandable by humans, and as a result, this lack of transparency and accountability can have severe consequences~\cite{rudin_stop_2019}, especially in safety-critical systems. 
However, model explainability is very important in a variety of engineering applications.

An \textit{interpretable} alternative to the ``black box" models and with considerably less assumptions is symbolic regression (SR). 
SR is a method for automatically finding a suitable algebraic expression that best describes the observed/sampled data \cite{koza1992genetic}. 
It is different from conventional regression techniques (e.g., linear regression, polynomial regression) in that SR does not rely on a specific a-priori model structure, but instead searches for the optimal model structure while simultaneously optimizing the model's parameters.  
The sole assumption made by SR is that the response surface \textit{can} be described algebraically~\cite{minnebo_empowering_nodate}. 
SR can be achieved by various methods, such as genetic programming (GP)~\cite{koza1992genetic,schmidt_distilling_nodate}, Bayesian methods~\cite{jin_bayesian_2020} and physics inspired artificial intelligence (AI)~\cite{udrescu_ai_2020}. 

Minimal human bias and low complexity of the modeling process
that allows function expressiveness and insights into the underlying data-generating process is of paramount importance. 
For aeronautical applications, safety is of the foremost significance and the consequences of failure or malfunction may be loss of life or serious injury, serious environmental damage, or harm to plant or property~\cite{lwears_rethinking_2012}.  
In aviation, properly understanding the data generating process can lead to developing  and improving existing physical models for nonlinear dynamical systems that could lead to new insights, as well as indicate faults and failures that can save lives and money in the context of prognostics and health management (PHM)~\cite{nguyen_review_nodate}.

Nowadays, with the growing generation of large amounts of data in the aviation industry (e.g. passing from snapshot to continuous data collection), many applications have been developed and improved. 
Some of them\footnote{Predix Platform: \url{https://www.ge.com/digital/iiot-platform} \\ IntelligentEngine: \url{https://www.rolls-royce.com/products-and-services/civil-aerospace/intelligentengine.aspx}\\ The MRO Lab - Prognos: \url{https://www.afiklmem.com/en/solutions/about-prognos}}  
are focused on engine health monitoring (EHM), as this is a central topic for engine manufacturers and operators.
Continuous engine operating data (CEOD) are collected at high frequencies in newer aircraft types, a development which -in combination with suitable algorithms- can improve the predictive capabilities for engine operators.
With the purpose of improving the availability and operability of assets, EHM monitors the state of individual engines or engine fleets, by making use of historical operational data, or of data generated during past events.
By optimizing maintenance operations, not only safety is improved, but also asset utilization can be optimized, leading to reduced costs and improved operational efficiency. 
This is an area of interest not only for engine operators and maintenance providers, but also for engine manufacturers.
The aim of these data-driven solutions is primarily to avoid imminent failures by identifying possible anomalies in the engine operation, and secondly, to prevent over-maintenance of parts and components, exploiting their full life span. 

From an operational context, the use of models like the one presented in this paper can assist engine users to understand in depth the evolution of the deterioration of their engines, while making more reliable predictions about the time for maintenance actions and mitigate the possible disruptions in their flight and passenger operations. 
At the same time, maintenance providers can predict the deterioration in detail and anticipate the physical state of the engines they will inspect and repair in the near future, without first having to wait for the real asset to be inducted in the shop. 
This way, they can streamline the maintenance process and provide more accurate quotations to their customers. 
Last, engine manufacturers -- apart from benefiting in their maintenance business, for the aforementioned reasons -- can use this type of work to understand in a better way the performance of their global fleet. 
This way, they can identify the influence of the different operating environments (e.g. presence of sand particles, salty water, air pollution, etc.) in the evolution of engine's health and incorporate their findings in the design of either newer versions of the same engines or even to future engine generations.

The temperature of the exhaust gases of an engine, known as Exhaust Gas Temperature (EGT) has evolved to become the standard industrial indicator of the health of an aircraft engine \cite{von2014review}.  
This is because it can capture the cumulative effect of deterioration in the isentropic efficiency of gas path components.

The above information motivates our main research question: Can we uncover a \textit{meaningful} algebraic relationship between the EGT and the other measured parameters present in the CEOD 
data, using SR? By \textit{meaningful} we mean that the analytic expression between the EGT and the other parameters should also be justifiable.
The longtime industrial standard in engine health monitoring is the analysis of static, snapshot data. This approach, despite being computationally lighter, cannot capture the dynamics of continuous engine operation during the different flight phases
Having said that, our main contribution in this work is the first, to our knowledge, attempt to use real-world, \textit{continuous} data collected along the entire flight duration at a recording frequency of $1$Hz in order to model analytically the EGT against the rest of the monitored flight parameters.
These data, termed continuous engine operational data (CEOD), allow for a more complete digital representation of the operational history of an engine.

The rest of this paper is organized as follows. In Section~\ref{sec:sr}, we give a brief introduction to SR and the framework used and in Section~\ref{sec:rel_work} we present related work done in the field of aviation and engineering in general.
In Section~\ref{sec:set_res}, we describe the data set used in this study, the experimental setup and the results.
Finally, in Section~\ref{sec:conclusion} we conclude this study, discuss limitations of the current work and suggest future research directions.

\section{Symbolic Regression}\label{sec:sr}

Symbolic regression (SR) is a methodology for finding a suitable algebraic expression that best describes the observed data \cite{koza1992genetic}. In symbolic regression no a-priori assumptions on the possible form of the expression is made, as in, for example, conventional regression models (e.g., linear regression). We could say that the latter class of models constrains the space of available expressions. The only assumption made by SR is that the relationship between the input and the output data \textit{can} be described analytically (or in a symbolic form) \cite{koza1992genetic}. In order to find the most appropriate solution, SR searches the space of mathematical expressions and estimates the corresponding parameters simultaneously~\cite{koza1992genetic,jin_bayesian_2020}. 

Performing this \textit{data-to-function} regression~\cite{koza1992genetic} is a sophisticated task. Various frameworks have been developed to tackle this problem, such as genetic programming (GP)~\cite{koza1992genetic,schmidt_distilling_nodate}, Bayesian methods~\cite{jin_bayesian_2020} and physics inspired artificial intelligence (AI)~\cite{udrescu_ai_2020}. 
In this work we use GP as our framework to perform SR on our data, as with the progress in the field of GP~\cite{koza1992genetic}, new ideas and methodologies have made GP a tool that could outperform more traditional techniques when solving modelling and identification problems, such us  autoregressive moving-average (ARMA) models \cite{wang2009comparison}. 
Furthermore GP provides a rather straightforward solution to the problem of SR.

\subsection{Genetic Programming}\label{sec:gp}

Genetic Programming (GP), first introduced by Koza~\cite{koza1992genetic} in $1992$, is a biologically inspired machine learning method that evolves computer programs to perform a specific task. 
When that task is building an empirical mathematical model then GP is called symbolic regression (SR). 
GP is a specialized form of genetic algorithms (GA)~\cite{forrest1993genetic}. 
GA is likely the most widely known type of Evolutionary Algorithms (EA) which comprise a larger class of direct, probabilistic search and optimization algorithms inspired from the model of organic structure evolution~\cite{holland_adaptation_1992,back_evolutionary_1996}.
The idea is to evolve randomly generated initial solutions (or chromosomes, as they are more commonly referred to) on a given problem following Darwin’s theory of evolution and to find the fittest solution after a number of generations or other user-specified termination criteria~\cite{forrest1993genetic}. 
Solution candidates are evolved through what are called genetic operators, which include crossover or recombination and mutation, as well as selection~\cite{back_evolutionary_1996,forrest1993genetic}.
Each individual solution is evaluated using a fitness function, which essentially tailors the evolutionary algorithm to the specific problem.
In essence, solutions are selected in a way that reflects their evaluation (better solutions have a higher chance of getting selected), recombined to make offspring solutions and in turn mutated, and replace the parent population for the next generation.
For more information on EA we refer the interested reader to~\cite{back_evolutionary_1996}.

\begin{figure}[hbt!]
    \centering
    \includegraphics[width = 0.2\textwidth]{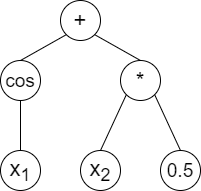}
    \caption{Basic GP tree representation.}
    \label{fig:GP_Tree}
\end{figure}

Instead of using strings of binary digits as chromosomes, to represent solutions, as in GA~\cite{forrest1993genetic}, solutions in GP are represented as tree-structured chromosomes, formed by nodes called operators and terminals.
As an example, Figure~\ref{fig:GP_Tree}
represents in a basic tree the simple expression:
\begin{equation}\label{eq:Eq1}
    (\cos(x_1) +  (x_2 \cdot 0.5))
\end{equation}
\myequations{Simple expression}

\begin{figure}[hbt!]
    \centering
    \includegraphics[width = 0.6\textwidth]{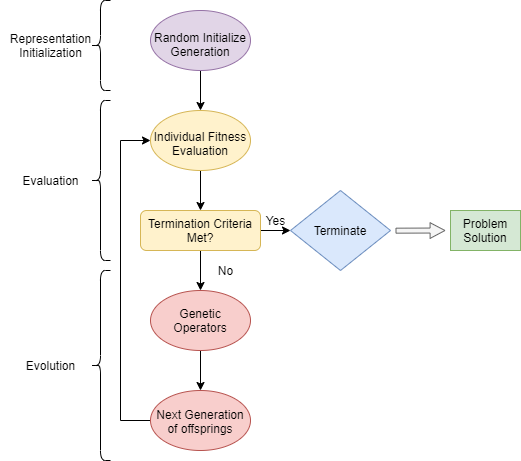}
    \caption{Genetic programming algorithm flowchart.}
    \label{fig:GP_Flowchart}
\end{figure}

Terminals are variables or values, that the operator can process. 
These include input variables like $x_i$ or coefficients to be used. 
The operators correspond to all those functions that can be applied to terminal nodes. 
These could be the fundamental arithmetic operators, such as $\{ +, -, \cdot, /, \exp, \log,  \sin, \cos, \ldots\}$, Boolean logic functions (AND, OR, NOT, etc) or any other mathematical functions. 
An individual (tree), is the hierarchical combination of operators and terminals, which is equivalent to an algebraic expression. 
When generating these tree structures, their computational complexity will be dependent on the used method for building them (hybrid, declarative, procedural, mathematical). 
A more detailed description of these tree building methodologies, as well as the algorithmic execution of a GP workflow can be found in \cite{sette2001genetic} and is illustrated in
Figure~\ref{fig:GP_Flowchart}.

The standard framework of GP, however, suffers from high complexity and overly complicated output expressions in SR~\cite{korns_accuracy_2011}. 
In order to mitigate these side effects, multi-gene genetic programming (MGGP) has been developed as a robust variant of GP~\cite{gandomi_new_2012}.
While the standard representation of a GP algorithm is based on the evaluation of one single tree structure,
MGGP is designed to generate individual members of the GP population (mathematical models of predictor
response  data) that are \textit{multi-gene} in nature, i.e., linear
combinations of low-order nonlinear transformations of
the input variables~\cite{searson2010gptips,gandomi_new_2012}. 
The user can specify the maximum allowable number of genes and the maximum tree depth any gene may have.
This facilitates a remarkable control over the maximum complexity of the evolved models~\cite{searson2010gptips,gandomi_new_2012}.

\begin{figure}[hbt!]
    \centering
    \includegraphics[width = 0.5\textwidth]{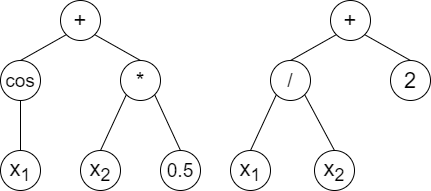}
    \caption{Multigene genetic programming model example.}
    \label{fig:Multi_model}
\end{figure}

Mathematically, a multi-gene regression model can be expressed as:
\begin{equation}\label{eq:Generic_Multi_model}
    \widehat{y} = d_0 + d_1 \cdot \text{Tree}_1 + \ldots + d_n \cdot \text{Tree}_n
\end{equation}
\myequations{Generic Multigene Regression model}
where \textit{$d_0$} represents the bias term, \textit{n} is the number of genes which constitutes a certain individual and $d_1,\ldots,d_n$
are the gene weights. 
Figure~\ref{fig:Multi_model} represents an example of a multigene genetic programming model that represents the mathematical expression in Equation~\ref{eq:MultigeneModel}:

\begin{equation}\label{eq:MultigeneModel}
    d_0 + d_1\cdot(\cos(x_1) +  (x_2\cdot0.5)) + d_2\cdot(x_1/x_2 + 2)
\end{equation}
\myequations{Multigene Regression model example}

\section{Related Work}\label{sec:rel_work}
Although, to the best of our knowledge, SR by means of GP has not been applied to the modelling of the EGT from real-life continuous flight data, there have been certain related studies.
A study closely resembling our work is from Nayyeri et al.~\cite{nayyeri_modeling_2012} who proposed an offline health monitoring system by simulating the EGT using SR by means of GP for the take-off and cruising phases of simulated data. 
The results returned an error of less than $0.5\%$ and $2.5\%$ for the take-off and cruising phases, respectively, indicating good performance.
However, the material used was simulated snapshot data and the authors did not use regularization to reduce model complexity.
Arellano et al.~\cite{arellano_prediction_2014} developed a SR approach by means of GP to predict future values of EGT, amongst other jet engine parameters, for control design. 
The data were collected from a small scale jet engine which operates on the same principles as the commercial jet engines. 
In \cite{li_linear--parameter_2006} the authors modeled the start-up process of an aero-engine, by performing SR using a specialized GP that generates models that are linear combinations of nonlinear functions of the inputs and produces more parsimonious solutions.
The main idea is to apply orthogonal least squares to estimate the contribution of the branches of the tree to the accuracy of the model.
The models outperform the results returned from the support vector machine (SVM) algorithm and in general can identify the dynamic system characteristics correctly, even without system knowledge.
GP has further been used in the field of aviation to nonlinear identification of aircraft engine~\cite{arkov_system_nodate,evans_application_2001,ruano_nonlinear_2003}.

There have also been numerous contributions of SR in engineering in general, apart from aviation. 
An example is \cite{enriquez_zarate_automatic_2016} were the authors used SR by means of a specially designed GP to predict the fuel flow and the EGT of a gas turbine in an electrical power setting.
Their approach outperformed machine-learning techniques and other symbolic regression techniques, such as fast function extraction (FFX) and multi-variate adaptive regression splines (MARS), on the EGT problem.
The results showed that standard GP algorithms can be used to address difficult real-world problems.
In \cite{togun_genetic_2010} the authors present the first approach for the formulation of a gasoline engine performance parameters (torque and brake specific fuel consumption) using an extension of GP called gene expression programming (GEP) that evolves computer programs encoded in linear chromosomes of fixed length. 
Their results demonstrate that GP can be effectively used to obtain formulations for high nonlinear function approximation problems in general.
Bongard and Lipson \cite{bongard2007automated} generated symbolic equations for nonlinear coupled dynamical systems in the fields of mechanics, systems biology and ecology. 
They also noted the differences between symbolic and numerical models in terms of complexity, making the former easier to interpret. 
In \cite{schmelzer_data-driven_2018} the authors developed a deterministic SR method to derive algebraic Reynolds-stress models for the Reynolds-averaged Navier-Stokes (RANS) equations, for turbulence modelling.

Genetic programming has also provided solutions to various problems such as classification problems \cite{zhang2004multiclass}, telecommunications problems \cite{faris2014genetic} and manufacturing process modelling \cite{faris2013modelling}.

The aforementioned list of applications is by no means exhaustive.
It shows, nevertheless, that GP can be successfully applied to real-world industrial problems, with better, comparable and intepretable results, compared to ``black box'' machine (ML) learning and artificial intelligence (AI) methods.
What is more, we can see that there is also a lot of potential and growing opportunities of GP applications in the field of aviation, still to come.
This work stands as an example of such an application on real-life turbo-fan engine data.

\section{Experimental Setup and Results}\label{sec:set_res}
Our objective is to see if the use of SR can uncover meaningful relationships in complex, engineering problems. 
Driven by this aim, we performed the following experiments on a real aircraft operational data set, in order to uncover relevant dependencies between the EGT and other measured parameters of a flight.

\subsection{Data} \label{sec:CEOD_Exp}

The data used in this study came from a specific GEnx turbofan engine, mounted on a Boeing $787-10$ and were recorded during four flights in July 2019\footnote{The data used is proprietary material of the Koninklijke Luchtvaart Maatschappij N.V. (KLM) and cannot be shared in the public domain.}.  
The collected data are termed continuous engine operational data (CEOD) \cite{forest2018generic} and are a data stream made out of several hundred parameters ($696$) which are measured along the entire flight duration at a recording frequency of $1$Hz.
Due to on-board computational limitations, the data have been off-loaded post-flight via gatelink.
The four different flights were anonymized for confidentiality and security purposes. 

An important point to be made is that the recording of CEOD is a relatively new technical development, so its use in engine health monitoring is still very limited from an operational standpoint. 
The longtime industrial standard is still the use of snapshot data, which are recorded only once during every flight phase. 
In other words, snapshot data contain only one point for takeoff and cruise, and, depending on the aircraft type, for the remaining flight phases. 
The advantage of CEOD for diagnostics and prognostics is obvious when combined with Machine Learning algorithms, since their training can be more effective. 

The selected target parameter that will be modelled is termed in the CEOD data set as \textit{Selected Exhaust Gas Temperature (DEG\_C)}. 
In the remainder of this study, we will call this simply \textit{EGT}.

\subsection{Experimental Setup}

\subsubsection{Data Pre-processing}

For this study, we decided to select the most stable phase of the flight, as exhibited by the data. 
This phase is assumed to be the cruising phase, due to the lack of labelled phase segmentation in the data.
This assumption was further validated by field experts.
This decision was made to allow for an accurate modelling of the underlying process, as the distribution of EGT measurements does not exhibit extreme fluctuations, since during cruise the operational and the environmental conditions are more stable compared to other flight phases.
Thus, phases such as, taxiing, take-off, climb, descent or landing were not investigated as they constitute a transient part of a flight, where engine performance and thermal effects vary with time and with mission characteristics.
Furthermore, the cruising phase allowed for a larger data sample, since it covers the greatest part of a long haul flight.
A large sample is important to uncover any meaningful relationships between the EGT and the rest of the monitored engine parameters.

For our experiments three of the flights, henceforth known as \textit{flight 1, flight 2 and flight 3}, were concatenated into a single data set.
From this data set we discarded parameters providing little to no information. 
Specifically, we removed parameters containing at least one NaN ($6.9$\% of the total parameters) value or string (alphanumeric), retaining only numeric data.
Subsequently, we split the remaining data into training and test sets by randomly selecting $80\%$ of the data for training and the remaining $20\%$ for testing. 
We further pre-processed the training data by performing a correlation analysis with different conditions that result in the different experiments (see Section \ref{sec:Methodology}). 
The training data will allow the SR algorithm to \textit{learn} patterns from the data and as a result estimate the model's parameters.
The test set is used in order to reduce any over-fitting of the SR algorithm to the training data, by estimating the generalization capability of the fitted model on the test data. 
This will reduce the possibility of the resulting algebraic expression reflecting \emph{only} the training data, from which it was generated.
The process of training and testing can be considered as the training phase of the SR algorithm.
A fourth flight (\textit{flight 4}) was selected for validation purposes in order to measure the final performance of our method on unseen data. 
For the validation and test data we only used the parameters that were retained on the training set after all the pre-processing steps performed on it. 
The final pre-processing step involved normalizing each of the parameters of the training, test and validation data as follows:
\begin{equation}
    x' = \frac{x-\mu}{\sigma},
\end{equation}
where $x, x'$ are the data item and the transformed data item, respectively and $\mu, \sigma$ are the population (or sample) mean and standard deviation, respectively.
It should be pointed out here that $\mu, \sigma$ which were used to normalize the validation and test parameters, are the same $\mu, \sigma$ learnt from the training data.
The latter step is standard practice in ML. 
Finally, we should note here that the selection of the flights in order to be used for training and testing has been done randomly, i.e., not taking into consideration flight details or characteristics, e.g.~departure airport or duration of the flight.
Furthermore, we would like to point out that there are different potential reasons for the presence of NaN values in the dataset. 
In general, NaN values can be attributed to recording and synchronization issues and to the fact that not all data capturing takes place at the exact same frequency, even if the \emph{recording} takes place at $1$Hz in the CEOD. In addition, not all parameters are recorded during all the phases of a flight, so a part of the missing values could be attributed to this reason. 
Moreover, some secondary systems might not be functional for operational reasons during specific segments or the totality of the flight. 
Moreover, temporary recording issues cannot be excluded. 
For calculated parameters, some required inputs might not be available at the time that certain entries are recorded, for the aforementioned reasons, resulting in NaN values. 
Finally, as a result of data ownership agreements between the original equipment manufacturer and the aircraft operator, some parameters are missing completely.

\subsubsection{Methodology and System Setup}\label{sec:Methodology}

We decided to use the \textit{GPTIPS}~\cite{searson2010gptips, searson_gptips_2015} to perform our experiments because of its ease of use, as well as its multigene GP approach that was discussed earlier.
Additionally, \textit{GPTIPS} takes into account the trade off surface of model performance and model complexity~\cite{searson2010gptips,searson_gptips_2015}.
In the multi-gene approach complexity is defined as the simple sum of the expressional complexities of its constituent trees~\cite{searson_gptips_2015}.
For \emph{each} experiment, $10$ final models were independently created, \emph{each} of which used $10$ independent runs \emph{internally}.
The models resulting from the multiple, \emph{internal} runs are automatically merged at the end of the execution and the best model is selected in terms of predictive performance ($\text{R}^2$ - see also Performance Metrics~\ref{sec:Error_Metrics} below) among models from a Pareto front of model performance and model complexity.
This internal, multi-start approach mitigates issues with the possible loss of model diversity over a single run and with the GP algorithm getting stuck in local minima~\cite{searson_gptips_2015}.
The repetition ($10$ times) of the previously mentioned process, per experiment, is performed to have an estimate of the centrality and dispersion of the performances in each experiment.  
The population size was chosen to be $250$ individuals, while the number of generations was at maximum $150$ generations. 
The tournament size is set to $20$, Tournament Pareto which encourages less complex models was set to $0.3$. 
Elitism = $30\%$ of the population. 
Maximum tree depth was set to $5$ and the maximum number of genes was selected to be $10$. 
Finally, the function set contained these operators = \{$\cdot, -, +, /, x^2, \sqrt, \exp, x^3, x^a, \exp(-x), -x, |x|, \log $\}.  
In essence these operators define our alphabet.
See Table \ref{tab:system_setup} for a quick reference of the hyperparameters used. 

\begin{table}[hbt!]
\centering
\begin{tabular}{l|c}
\multicolumn{1}{c|}{\textbf{Hyperparameter}} &
  \textbf{Value} \\ \hline
Runs (internal)                    & 10  \\
Population size         & 250 \\
Number of generations   & 150 \\
Tournament size         & 20  \\
Tournament Pareto       & 0.3 \\
Elitism                 & 0.3 \\
Maximum tree depth      & 5   \\
Maximum number of genes & 10  \\
Function set & \begin{tabular}[c]{@{}c@{}}$\cdot, -, +, /, x^2, \sqrt, \exp, x^3, x^a, \exp(-x), -x, |x|, \log $\end{tabular} \\ \hline
\end{tabular}
\caption{System setup parameters.}
\label{tab:system_setup}
\end{table}

By default \textit{GPTIPS} provides a multigene symbolic regression fitness function, which was used in order to minimize the root mean squared prediction error on the training data. 
For the genetic operators we used $p=0.1$ for the mutation probability and $p=0.85$ for the crossover probability.
The chosen hyperparameters were based on values suggested from literature, in combination with execution time, and preliminary experiments. More specifically, the mutation/crossover rates are equal to the values in~\cite{searson2010gptips} (default values). 
The same is also true for other non-mentioned hyperparameters of the algorithm.

\subsubsection{Performance Metrics}\label{sec:Error_Metrics}

To measure the performance of our approach against the ground truth, we decided to use the common error metrics for regression~\cite{shcherbakov2013survey}, namely, root mean squared error (RMSE), mean squared error (MSE), mean absolute error (MAE), and R$^2$:
\begin{multicols}{2}
    \begin{equation}\label{eq:RMSE}
        RMSE(y,\hat{y}) = \sqrt{\frac{1}{n}\Sigma_{i=1}^{n}{({y_i -\hat{y}_i)}^2}}
    \end{equation}

     \begin{equation}\label{eq:MSE}
        MSE(y,\hat{y}) = \frac{1}{n}\Sigma_{i=1}^{n}{({y_i -\hat{y}_i)}^2}
    \end{equation}
\end{multicols}
\begin{multicols}{2}
 \begin{equation}\label{eq:MAE}
        MAE(y,\hat{y}) = \frac{1}{n}\sum_{i=1}^{n}|y_i - \hat{y}_i|
    \end{equation}

    \begin{equation}\label{eq:R2}
        R^2(y,\hat{y}) = 1 - \frac{\sum_{i}^{n}(y_i - \hat{y}_i)^2}{\sum_{i}^{n}(y_i -\bar{y})^2}  
    \end{equation}
    
\end{multicols}

where \textit{$y_i$} are the ground truth values, $\hat{y}_i$ the predicted values, $\bar{y}$ is the mean of the observed data, and \textit{$n$} the number of samples.

\subsection{Experimental Results}\label{sec:results}

All experiments were executed on an off-the-shelf PC with a processor running at $1.8$ GHz and $8$ GB of RAM.
The source code has been developed using \textit{Python} version $3.8.3$ and \textit{MATLAB} version R$2019$b.
We used \textit{GPTIPS} version $2$ and \textit{Pandas} version $1.0.3$. 

For the experiment a correlation analysis was performed on the input parameters, as a dimensionality reduction step. Specifically, parameters that were highly correlated (over $0.90$) were discarded, retaining only the first representative. 
After this step, $114/696$ CEOD parameters remained to be used as final input, in the GP framework, in addition to the target EGT.
We performed this experiment $10$ times to account for the stochastic nature of GP, by combining internally in each of these executions, $10$ independent runs. 
This resulted in $10$ different algebraic expressions.
In Table~\ref{tab:Exp1_error_metris} we show the average error metrics (over $10$ runs) for the training, the test, and the validation data sets.
The results show us that SR has managed to account for the variability of the EGT against the used CEOD parameters on both the training and test sets ($R^2=1)$. 
As a result, the average deviation from the ground truth is less than $1$ degree Celcius  (MAE= $0.77$\textdegree C and $0.76$\textdegree C) 
for the training and test sets, respectively, which is negligible from an engineering perspective.
Regarding the validation set, we see a larger average error, compared to the training and test sets. 
In particular, we see an average error of $3$\textdegree degrees Celcius compared to the ground truth EGT values.
This slight increase in the error, is also backed up by the small decrease of the average $R^2=0.86$, indicating a small degree of overfitting.
The small error increase is in general expected.
Taking into consideration, however, that we did not correct for parameters such as the duration of the cruising phase, or the flight level, the current error is within an acceptable range.

\begin{table}[hbt!]
\centering
\begin{tabular}{l|l|l|l|l|}
\cline{2-5}
& $R^2$ & RMSE & MAE  & MSE   \\ \hline
\multicolumn{1}{|l|}{Training Error}   & 1 $\pm$ 0                   & 1.3 $\pm$ 0.06  & 0.77 $\pm$ 0.03 & 1.69 $\pm$ 0.16  \\ \hline
\multicolumn{1}{|l|}{Test error} & 1 $\pm$ 0 & 1.27 $\pm$ 0.05 & 0.76 $\pm$ 0.03 & 1.62 $\pm$ 0.15 \\ \hline
\multicolumn{1}{|l|}{Validation error} & 0.86 $\pm$ 0.08    & 8.44 $\pm$ 3.27 & 3.01 $\pm$ 1.01 & 80.81 $\pm$ 46.8 \\ \hline
\end{tabular}
\caption{Average error metrics (over $10$ runs) of the training error, the test error and the validation error, on experiment $1$.}
\label{tab:Exp1_error_metris}
\end{table}

\begin{figure}[hbt!]
    \centering
    \includegraphics[width=11cm]{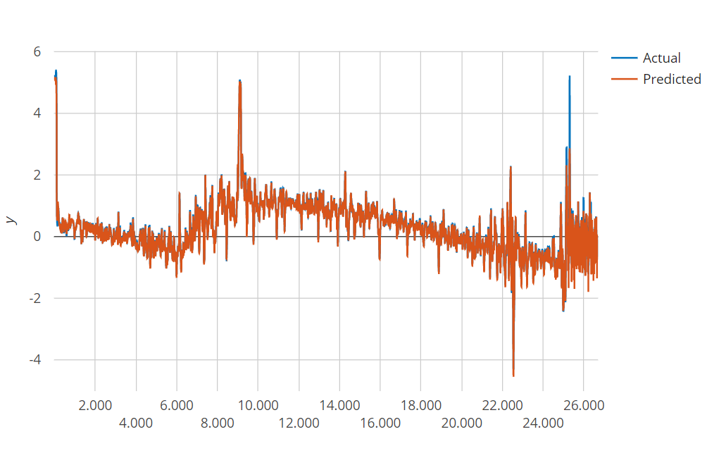}
    \caption{Scaled EGT predictions (red) ($y$-axis) on the validation set 
    vs.~observed (blue) values (Model 1 - Experiment 1). $x$-axis shows the data sample index in consecutive order according to their sampling over time.}
    \label{fig:Exp1_model1}
\end{figure}

In Figure~\ref{fig:Exp1_model1} we show a plot of the EGT predictions on the validation set (displayed in orange) overlaid against the observed EGT values (displayed in blue). 
The $x$-axis represents the number of used data-points. 
We should note here that the \textit{y} axis represents the \textit{scaled} EGT measures. 
From the results we can see that the resulting algebraic expression has managed to learn the underlying relationship between the EGT and the other CEOD parameters very well.
In addition, in Table~\ref{tab:Exp1_error_metris} we can see that the dispersion of the predictions against the observed EGT is, on average, $3$\textdegree C.

In addition, Equation~\ref{eq:Y1_1} is the algebraic expression for the first of the $10$ resulting models. 
\begin{equation}\label{eq:Y1_1}
\begin{split}
    Y_{1}^{1}  & = 0.141\cdot x_{4} + 0.123\cdot x_{5} + 0.8\cdot x_{6} + 0.0214\cdot x_{12} - 0.123\cdot x_{18} + 0.751\cdot x_{21} + 0.0261\cdot x_{60} \\
     & + 0.0405\cdot x_{74} - 0.0371\cdot |\log(x_{39})| + 1.32\cdot10^{-4}\cdot e^{(2\cdot x_{21})} - 0.0428\cdot |x_{4}| \\
     & - 0.0261\cdot e^{(x_{21})} + 0.00762\cdot x^2_{74} + 0.133 
\end{split}
\end{equation}
\myequations{Model 1 of Experiment 1}
In Table~\ref{tab:percentage_app_exp1}, we show the input variables that appear in the resulting $10$ models as well as their percentage of appearance.
Here, each variable is represented by an $x$ with an index. 
\begin{table}[hbt!]
\centering
\resizebox{8cm}{!}{
\begin{tabular}{l|c}
\textbf{Input variables (Index)} & \textbf{\% of appearance} \\ \hline
$x_4$,   $x_6$, $x_{21}$                                     & 10,6 \%                                 \\
$x_{43}$                                                     & 8,5 \%                                  \\
$x_{12}$,   $x_{74}$, $x_{113}$                              & 5,1 \%                                  \\
$x_5$,   $x_{110}$                                           & 4,08 \%                                 \\
$x_{11}$,   $x_{14}$, $x_{39}$, $x_{59}$                     & 3,06 \%                                 \\
$x_{18}$,   $x_{23}$, $x_{24}$, $x_{94}$                     & 2,04 \%                                 \\
\begin{tabular}[c]{@{}l@{}}$x_1$, $x_8$, $x_{13}$, $x_{25}$, $x_{46}$, $x_{52}$,\\  $x_{57}$, $x_{60}$, $x_{96}$, $x_{111}$, $x_{112}$\end{tabular} & 1,02 \% \\ \hline
\end{tabular}} 
\caption{Percentage of appearance per variable over all models (Experiment 1).}
\label{tab:percentage_app_exp1}
\end{table}
The reader might find interesting to know which of the variables have resulted from this experiment. 
In the following list, we provide the technical meaning of the most frequently occurring variables based on Table~\ref{tab:percentage_app_exp1}. In the technical explanations, we considered only parameters with more than 5\% occurrence in Table~\ref{tab:percentage_app_exp1}.

\begin{itemize}
    \item \textbf{Actual Calculated HPT Clearance – $x_4$} \\
    The tip clearance of the High Pressure Turbine (HPT) is directly related to its isentropic efficiency and the gas enthalpy drop through the blade stages. 
    The higher the clearance, the less efficient the expansion process is, and thus the EGT is higher.
    \item \textbf{Average Gas Temperature at Station 25 – $x_6$} \\
    This is the gas temperature at the inlet of the High Pressure Compressor (HPC). 
    A higher temperature here indicates a less efficient compression process through the engine booster, which for a given pressure ratio requires increased power input from its corresponding turbine, the Low Pressure Turbine (LPT). 
    This high power output can only be achieved via higher fuel flows that lead to increased EGT.
    \item \textbf{Corrected Fan Speed to Station 12 – $x_{21}$} \\
    A higher Fan Speed also corresponds to a higher EGT, since the power required from the interconnected LPT is higher, leading to an increased fuel flow.
    \item \textbf{FSV Minimum Main Fuel Split Regulator – $x_{43}$} \\
    As the Fuel Splitting Valve (FSV) has an influence on the amount of fuel that is directed to the combustion chamber, there is a direct relation between this variable and the resulting EGT.
    \item \textbf{BPCU 1 GCU Generator Load – $x_{12}$} \\
    This parameter is related to the load control of the engine generators. 
    The higher the load required from the generators, the higher the power extraction from the engine, which leads to higher fuel flow, to cover the increased energy needs. 
    The higher fuel flow results in a higher EGT.
    \item \textbf{Selected Variable Bleed Valve (VBV) Position – $x_{74}$} \\
    The position of the VBV controls the amount of air that is bled from the engine. 
    With an increasing degree of bleed, the HPC compresses air that does not contribute to the power generated by the turbines, resulting in a reduced overall thermal efficiency. 
    This reduction means that for the same thrust output, the engine needs to consume a higher amount of fuel, which results in an increased EGT.
    \item \textbf{WF$/$(P3 $\cdot$ RTH25) Base (PPH$/$PSIA) – $x_{113}$}\\
    This is an expression for the non-dimensionalized fuel flow of the engine, which is directly related to a higher EGT.
\end{itemize}

In the list above, the station numbering (e.g., ``Average Gas Temperature at Station 25'') is standardized and follows the SAE Aerospace Standard AS755 (Aircraft Propulsion System Performance Station Designation)\footnote{\url{https://www.sae.org/standards/content/as755g/}}. 
Under this standard, station 25 (see ``Average Gas Temperature at Station 25'') is the interface between the Low Pressure Compressor (LPC) and the HPC, while station 12 (see ``Corrected Fan Speed to Station 12'') is the inlet fan tip station.

Moreover, the coefficients multiplied by the variables in Equation~\ref{eq:Y1_1} indicate the relative importance (contribution) of that parameter to the output.
For example, the coefficients of variables $x_4, x_6,x_{21}$ are the largest among the coefficients of the other variables, showing their importance to the EGT.
This is also backed up by the percentage of appearance of these variables throughout the repetitions, as well as from the nature of these variables as mentioned before.

In addition, we performed two control experiments to investigate the effect that certain parameters might have on the estimation of the EGT.
In particular, in the first we removed the parameter \textit{Average Temperature at Station 25 (DEG\_C)}, which even though did not exceed the $0.9$ correlation threshold, is in direct relation with the EGT.
After removing it we performed the same experiment described before which resulted in Table~\ref{tab:Exp2_error_metris}.
The results show a similar pattern to those of our initial experiments.
In addition, we see a slight decrease (by $6\%$) in the average $R^2$ value of the validation set and a small increase (by $8\%$) in the average MAE value.
Despite the error increase, the deviation from the ground truth is still minimal, indicating that the EGT can be evaluated non-trivially. 
With this we mean that despite dropping parameters that are closely related to the nature of our target output (e.g., \textit{Average Temperature at Station 25 (DEG\_C)}), we still get satisfying results.

\begin{table}[hbt!]
\centering
\begin{tabular}{l|l|l|l|l|}
\cline{2-5}
& $R^2$ & RMSE & MAE  & MSE   \\ \hline
\multicolumn{1}{|l|}{Training Error}   & 1 $\pm$ 0                   & 1.43 $\pm$ 0.06  & 0.88 $\pm$ 0.03 & 2.03 $\pm$ 0.17  \\ \hline
\multicolumn{1}{|l|}{Test error} & 1 $\pm$ 0 & 1.4 $\pm$ 0.07 & 0.87 $\pm$ 0.04 & 1.97 $\pm$ 0.2 \\ \hline
\multicolumn{1}{|l|}{Validation error} & 0.81 $\pm$ 0.04    & 10.27 $\pm$ 1.24 & 3.26 $\pm$ 0.26 & 106.93 $\pm$ 23.73 \\ \hline
\end{tabular}
\caption{Average error metrics (over $10$ runs) of the training error, the test error and the validation error, on experiment $2$.}
\label{tab:Exp2_error_metris}
\end{table}

The second experiment involved discarding all the highly correlated (more than $90\%$) input parameters with the EGT before performing the correlation analysis of our initial experiment.
The results of this experiment is summarized in Table~\ref{tab:Exp3_error_metris}.
Here we see the results resembling more closely those of our initial experiment.
It is interesting, however, to see a decrease (by $7\%$) of the average MAE.

\begin{table}[hbt!]
\centering
\begin{tabular}{l|l|l|l|l|}
\cline{2-5}
& $R^2$ & RMSE & MAE  & MSE   \\ \hline
\multicolumn{1}{|l|}{Training Error}   & 1 $\pm$ 0                   & 1.23 $\pm$ 0.04  & 0.75 $\pm$ 0.02 & 1.5 $\pm$ 0.1  \\ \hline
\multicolumn{1}{|l|}{Test error} & 1 $\pm$ 0 & 1.24 $\pm$ 0.04 & 0.75 $\pm$ 0.02 & 1.55 $\pm$ 0.09 \\ \hline
\multicolumn{1}{|l|}{Validation error} & 0.86 $\pm$ 0.08    & 8.44 $\pm$ 3.23 & 2.8 $\pm$ 0.95 & 80.65 $\pm$ 44.63 \\ \hline
\end{tabular}
\caption{Average error metrics (over $10$ runs) of the training error, the test error and the validation error, on experiment $3$.}
\label{tab:Exp3_error_metris}
\end{table}

The resulting models and plots from all experiments can be found in the supplemental material accompanying this paper.

\section{Discussion and Future Work}\label{sec:conclusion}

In this work, we investigated the use of symbolic regression (SR) by means of genetic programming (GP) on a real engineering problem.
Specifically, we examined the use of SR on real aircraft operational data with the aim of uncovering meaningful relationships between the exhaust gas temperature (EGT) - a standard industrial indicator of the health of an aircraft engine - and the rest of the monitored engine parameters.
Our main contribution is the first, to our knowledge, analytical model of EGT against the rest of the monitored flight parameters which has been automatically derived from real-world \textit{continuous} data collected along the entire flight duration at a recording frequency of $1$Hz (and been assessed by engine experts to provide useful insights).
These data, termed continuous engine operational data (CEOD), allow for a more complete digital representation of the operational history of an engine, while the longtime industrial standard is still the use of snapshot data, which are recorded only once during every flight phase.

The experimental results are promising, both in terms of model accuracy, as well as in explainability.
In more detail, the trained models exhibited on average a small amount of overfitting and an absolute difference of $3$\textdegree C compared to the ground truth EGT values, a small difference from an engineering perspective.
Furthermore, the resulting formulas demonstrated consistency from a physics/engineering point of view between the predictor-parameters and the EGT, which was validated by field experts.
This indicated that the proposed method can uncover meaningful relationships in the data that can be interpreted by the end-user.
In addition, we performed two more experiments to investigate the effect that certain parameters might have on the estimation of the EGT.
The results showed a similar pattern to our initial experimental output.

The importance of our study lies in the fact that with little or no field knowledge, we were able to generate models that relate the EGT accurately and meaningfully to other monitored parameters.
Such algebraic expressions can assist field practitioners in diagnosing faults or failures and can even uncover new relationships between parameters, previously unknown to engineers or field experts.

At this point, we should also mention some of the limitations of our work.
Firstly, we only took into account the cruising phase of the flight, ignoring the others.
Having said that we expect a different behavior in phases such as take-off, where the engine performance is transient and thermally unstable.
Moreover, we did not take into account or correct in any way the data, based on information such as the cruising flight-level (altitude) or the duration of the flight or the weight of the aircraft during cruising.
For example, EGT might increase with increasing HPT tip clearance, since its isentropic thermal efficiency drops.
Moreover, even though the data pre-processing that we did proved to be effective, it meant we had to discard certain data because of the NaN values.
Lastly, we only modeled the EGT as a function of the rest of the observed parameters.
Modeling other parameters might be more difficult or even impossible.
However, as EGT is the standard industrial indicator for the overall engine thermal efficiency, this is not a main concern.

Our aforementioned limitations clearly pave the road into future directions.
Initially, we would like to model transient flight phases, such as take-off, which constitutes a very intensive time for the engine.
Additionally, it is worth looking into pre-processing the data with minimum loss of information (e.g., NaN value imputation) and to incorporate additional information (data augmentation), such as weather conditions (e.g., when modeling parameters during climb or landing).
Regarding the CEOD data specifically, we should emphasize that they can play a significant role since their higher sampling rate can capture, for example, early issues and pinpoint the exact moment they took place. 
However, since CEOD contain a larger amount of information compared to e.g., snapshot data, this means that the amount of data for training and testing need to be equally high. 
In addition, as the engine conditions might vary significantly during take off, due to different ambient conditions, airport elevation, engine derate, etc., data representativeness is a key-point for a successful application of the methods we used, and any other ML method in essence.
Regarding the modeling, it would very interesting to perform hyperparameter optimization on the GP in order to select the optimal hyperparameters that will allow high accuracy and low generalization error. It would also be of worth, to further build a meta-model that combines all of the formulas derived from the experiments or an ensemble model by, for example, taking the average or other aggregation function of the predictions provided by each of the models.
Also, like mentioned before, such models that are intepretable by the end-users, can lend themselves for predictive maintenance.
For example, any strong deviation between the predicted value of the model and the monitored parameter(s) can indicate a fault or malfunctioning sensor.
This of course would be possible if the model is built from \textit{healthy} data.
These formulas can also be used in order to generate more data healthy or faulty, by tuning the range of the predictor parameters to simulate various conditions.
Lastly, by proper data pre-processing, one can also derive formulas that allow forecasting of parameters into the future enabling this way prognostics.
This list is by no means exhaustive, but it is clear that the opportunities are endless.

\section*{Funding Sources}
This work is part of the research programme Smart Industry SI2016 with project name CIMPLO and project number 15465, which is partly financed by the Netherlands Organisation for Scientific Research (NWO).

\bibliography{mybibliography}

\begin{thebibliography}{36}
\newcommand{\enquote}[1]{``#1''}
\providecommand{\natexlab}[1]{#1}
\providecommand{\url}[1]{\texttt{#1}}
\providecommand{\urlprefix}{URL }
\expandafter\ifx\csname urlstyle\endcsname\relax
  \providecommand{\doi}[1]{\discretionary{}{}{}https://doi.org/#1}\else
  \providecommand{\doi}[1]{\discretionary{}{}{}\urlstyle{rm}\url{https://doi.org/#1}}\fi

\bibitem[{Vaddireddy et~al.(2020)Vaddireddy, Rasheed, Staples, and
  San}]{vaddireddy_feature_2020}
Vaddireddy, H., Rasheed, A., Staples, A.~E., and San, O., \enquote{Feature
  Engineering and Symbolic Regression Methods for Detecting Hidden Physics from
  Sparse Sensors,} \emph{Physics of Fluids}, Vol.~32, No.~1, 2020, p. 015113.
\newblock \doi{10.1063/1.5136351},
  \urlprefix\url{http://arxiv.org/abs/1911.05254}, arXiv: 1911.05254.

\bibitem[{Box et~al.(2015)Box, Jenkins, Reinsel, and Ljung}]{box2015time}
Box, G.~E., Jenkins, G.~M., Reinsel, G.~C., and Ljung, G.~M., \emph{Time Series
  Analysis: Forecasting and Control}, John Wiley \& Sons, 2015.

\bibitem[{{Ian Goodfellow} et~al.(2016){Ian Goodfellow}, {Yoshua Bengio}, and
  {Aaron Courville}}]{ian_goodfellow_deep_2016}
{Ian Goodfellow}, {Yoshua Bengio}, and {Aaron Courville}, \emph{Deep
  {Learning}}, MIT Press, 2016.
\newblock \urlprefix\url{http://www.deeplearningbook.org}.

\bibitem[{Rudin(2019)}]{rudin_stop_2019}
Rudin, C., \enquote{Stop {Explaining} {Black} {Box} {Machine} {Learning}
  {Models} for {High} {Stakes} {Decisions} and {Use} {Interpretable} {Models}
  {Instead},} \emph{arXiv:1811.10154 [cs, stat]}, 2019.
\newblock \urlprefix\url{http://arxiv.org/abs/1811.10154}, arXiv: 1811.10154.

\bibitem[{Koza(1992)}]{koza1992genetic}
Koza, J.~R., \emph{Genetic Programming: on the Programming of Computers by
  Means of Natural Selection}, Vol.~1, MIT press, 1992.

\bibitem[{Minnebo and Stijven(2011)}]{minnebo_empowering_nodate}
Minnebo, W., and Stijven, S., \enquote{Empowering {Knowledge} {Computing} with
  {Variable} {Selection} - {On} {Variable} {Importance} and {Variable}
  {Selection} in {Regression} {Random} {Forests} and {Symbolic} {Regression},}
  Master's thesis, University of Antwerp, Antwerp, Belgium, 2011.

\bibitem[{Schmidt and Lipson(2009)}]{schmidt_distilling_nodate}
Schmidt, M., and Lipson, H., \enquote{Distilling Free-Form Natural Laws From
  Experimental Data,} \emph{Science}, Vol. 324, No. 5923, 2009, pp. 81--85.
\newblock \doi{https://doi.org/10.1126/science.1165893}.

\bibitem[{Jin et~al.(2020)Jin, Fu, Kang, Guo, and Guo}]{jin_bayesian_2020}
Jin, Y., Fu, W., Kang, J., Guo, J., and Guo, J., \enquote{Bayesian {Symbolic}
  {Regression},} \emph{arXiv:1910.08892 [stat]}, 2020.
\newblock \urlprefix\url{http://arxiv.org/abs/1910.08892}, arXiv: 1910.08892.

\bibitem[{Udrescu and Tegmark(2020)}]{udrescu_ai_2020}
Udrescu, S.-M., and Tegmark, M., \enquote{{AI} {Feynman}: {A} Physics-Inspired
  Method for Symbolic Regression,} \emph{Science Advances}, Vol.~6, No.~16,
  2020, p.~16.
\newblock \doi{10.1126/sciadv.aay2631},
  \urlprefix\url{https://www.ncbi.nlm.nih.gov/pmc/articles/PMC7159912/}.

\bibitem[{Lwears(2012)}]{lwears_rethinking_2012}
Lwears, R., \enquote{Rethinking Healthcare as a Safety--Critical Industry,}
  \emph{Work (Reading, Mass.)}, Vol. 41 Suppl 1, 2012, pp. 4560--4563.
\newblock \doi{10.3233/WOR-2012-0037-4560}.

\bibitem[{Nguyen et~al.(2019)Nguyen, Kefalas, Yang, Apostolidis, Olhofer, and
  Limmer}]{nguyen_review_nodate}
Nguyen, V.~D., Kefalas, M., Yang, K., Apostolidis, A., Olhofer, M., and Limmer,
  S., \enquote{A {Review}: {Prognostics} and {Health} {Management} in
  {Automotive} and {Aerospace},} \emph{International Journal of Prognostics and
  Health Management}, Vol.~10, No.~2, 2019, p.~35.
\newblock \doi{https://doi.org/10.36001/ijphm.2019.v10i2.2730}.

\bibitem[{Von~Moll et~al.(2014)Von~Moll, Behbahani, Fralick, Wrbanek, and
  Hunter}]{von2014review}
Von~Moll, A., Behbahani, A.~R., Fralick, G.~C., Wrbanek, J.~D., and Hunter,
  G.~W., \enquote{A Review of Exhaust Gas Temperature Sensing Techniques for
  Modern Turbine Engine Controls,} \emph{50th AIAA/ASME/SAE/ASEE Joint
  Propulsion Conference}, Cleveland, OH, USA, 2014, pp. 2014--3977.
\newblock \doi{https://doi.org/10.2514/6.2014-3977}.

\bibitem[{Wang et~al.(2009)Wang, Chau, Cheng, and Qiu}]{wang2009comparison}
Wang, W.-C., Chau, K.-W., Cheng, C.-T., and Qiu, L., \enquote{A Comparison of
  Performance of Several Artificial Intelligence Methods for Forecasting
  Monthly Discharge Time Series,} \emph{Journal of hydrology}, Vol. 374, No.
  3-4, 2009, pp. 294--306.
\newblock \doi{https://doi.org/10.1016/j.jhydrol.2009.06.019}.

\bibitem[{Forrest(1993)}]{forrest1993genetic}
Forrest, S., \enquote{Genetic Algorithms: Principles of Natural Selection
  Applied to Computation,} \emph{Science}, Vol. 261, No. 5123, 1993, pp.
  872--878.
\newblock \doi{https://doi.org/10.1126/science.8346439}.

\bibitem[{Holland(1992)}]{holland_adaptation_1992}
Holland, J. H. J.~H., \emph{Adaptation in {Natural} and {Artificial} Systems :
  an Introductory Analysis with Applications to Biology, Control, and
  Artificial Intelligence}, Cambridge, Mass. : MIT Press, 1992.
\newblock \urlprefix\url{http://archive.org/details/adaptationinnatu00holl}.

\bibitem[{Bäck(1996)}]{back_evolutionary_1996}
Bäck, T., \emph{Evolutionary Algorithms in Theory and Practice: Evolution
  Strategies, Evolutionary Programming, Genetic Algorithms}, Oxford University
  Press, Inc., USA, 1996.

\bibitem[{Sette and Boullart(2001)}]{sette2001genetic}
Sette, S., and Boullart, L., \enquote{Genetic Programming: Principles and
  Applications,} \emph{Engineering Applications of Artificial Intelligence},
  Vol.~14, No.~6, 2001, pp. 727--736.
\newblock \doi{https://doi.org/10.1016/S0952-1976(02)00013-1}.

\bibitem[{Korns(2011)}]{korns_accuracy_2011}
Korns, M.~F., \enquote{Accuracy in {Symbolic} {Regression},} \emph{Genetic
  {Programming} {Theory} and {Practice} {IX}}, edited by R.~Riolo,
  E.~Vladislavleva, and J.~H. Moore, Genetic and {Evolutionary} {Computation},
  Springer, New York, NY, 2011, pp. 129--151.
\newblock \doi{10.1007/978-1-4614-1770-5_8},
  \urlprefix\url{https://doi.org/10.1007/978-1-4614-1770-5_8}.

\bibitem[{Gandomi and Alavi(2012)}]{gandomi_new_2012}
Gandomi, A.~H., and Alavi, A.~H., \enquote{A New Multi-Gene Genetic Programming
  Approach to Nonlinear System Modeling. {Part} {I}: Materials and Structural
  Engineering Problems,} \emph{Neural Computing and Applications}, Vol.~21,
  No.~1, 2012, pp. 171--187.
\newblock \doi{10.1007/s00521-011-0734-z},
  \urlprefix\url{https://doi.org/10.1007/s00521-011-0734-z}.

\bibitem[{Searson et~al.(2010)Searson, Leahy, and Willis}]{searson2010gptips}
Searson, D.~P., Leahy, D.~E., and Willis, M.~J., \enquote{GPTIPS: An Open
  Source Genetic Programming Toolbox for Multigene Symbolic Regression,}
  \emph{Proceedings of the International multiconference of engineers and
  computer scientists}, Vol.~1, Newswood Ltd., Hong Kong, 2010, pp. 77--80.

\bibitem[{Nayyeri and Khorasani(2012)}]{nayyeri_modeling_2012}
Nayyeri, H., and Khorasani, K., \enquote{Modeling Aircraft Jet Engine and
  System Identification by Using {Genetic} {Programming},} \emph{2012 25th
  {IEEE} {Canadian} {Conference} on {Electrical} and {Computer} {Engineering}
  ({CCECE})}, IEEE, Montreal, QC, 2012, pp. 1--4.
\newblock \doi{10.1109/CCECE.2012.6334869},
  \urlprefix\url{http://ieeexplore.ieee.org/document/6334869/}.

\bibitem[{Arellano et~al.(2014)Arellano, Cant, and
  Nolle}]{arellano_prediction_2014}
Arellano, G.~M., Cant, R., and Nolle, L., \enquote{Prediction of {Jet} {Engine}
  {Parameters} for {Control} {Design} {Using} {Genetic} {Programming},}
  \emph{2014 {UKSim}-{AMSS} 16th {International} {Conference} on {Computer}
  {Modelling} and {Simulation}}, IEEE, Cambridge, United Kingdom, 2014, pp.
  45--50.
\newblock \doi{10.1109/UKSim.2014.64},
  \urlprefix\url{http://ieeexplore.ieee.org/document/7046037/}.

\bibitem[{Li and Wei(2006)}]{li_linear--parameter_2006}
Li, Y.-h., and Wei, X.-k., \enquote{Linear-in-{Parameter} {Models} {Based} on
  {Parsimonious} {Genetic} {Programming} {Algorithm} and {Its} {Application} to
  {Aero}-{Engine} {Start} {Modeling},} \emph{Chinese Journal of Aeronautics},
  Vol.~19, No.~4, 2006, pp. 295--303.
\newblock \doi{10.1016/S1000-9361(11)60331-2},
  \urlprefix\url{https://linkinghub.elsevier.com/retrieve/pii/S1000936111603312}.

\bibitem[{Arkov et~al.(2000)Arkov, Evans, Fleming, and
  Hill}]{arkov_system_nodate}
Arkov, V., Evans, C., Fleming, P.~J., and Hill, D.~C., \enquote{{System}
  {Identification} {Strategies} {Applied} {to} {Aircraft} {Gas} {Turbine}
  {Engines},} \emph{Annual Reviews in Control}, Vol.~24, 2000, pp. 67--81.
\newblock \doi{https://doi.org/10.1016/S1367-5788(00)90015-4}.

\bibitem[{Evans et~al.(2001)Evans, Fleming, Hill, Norton, Pratt, Rees, and
  Rodr}]{evans_application_2001}
Evans, C., Fleming, P.~J., Hill, D.~C., Norton, J.~P., Pratt, I., Rees, D., and
  Rodr, K., \enquote{Application of System Identification Techniques to
  Aircraft Gas Turbine Engines,} \emph{Control Engineering Practice}, Vol.~9,
  No.~2, 2001, p.~14.
\newblock \doi{https://doi.org/10.1016/S0967-0661(00)00091-5}.

\bibitem[{Ruano et~al.(2003)Ruano, Fleming, Teixeira, Rodriguez-Vazquez, and
  Fonseca}]{ruano_nonlinear_2003}
Ruano, A., Fleming, P., Teixeira, C., Rodriguez-Vazquez, K., and Fonseca, C.,
  \enquote{Nonlinear identification of aircraft gas-turbine dynamics,}
  \emph{Neurocomputing}, Vol.~55, No. 3-4, 2003, pp. 551--579.
\newblock \doi{10.1016/S0925-2312(03)00393-X},
  \urlprefix\url{https://linkinghub.elsevier.com/retrieve/pii/S092523120300393X}.

\bibitem[{Enriquez\_Zárate et~al.(2017)Enriquez\_Zárate, Trujillo, De~Lara,
  Castelli, Z-Flores, Muñoz, and Popovič}]{enriquez_zarate_automatic_2016}
Enriquez\_Zárate, J., Trujillo, L., De~Lara, S., Castelli, M., Z-Flores, E.,
  Muñoz, L., and Popovič, A., \enquote{Automatic {Modeling} of a {Gas}
  {Turbine} using {Genetic} {Programming}: {An} {Experimental} {Study},}
  \emph{Applied Soft Computing}, Vol.~50, 2017, pp. 212--222.
\newblock \doi{10.1016/j.asoc.2016.11.019}.

\bibitem[{Togun and Baysec(2010)}]{togun_genetic_2010}
Togun, N., and Baysec, S., \enquote{Genetic Programming Approach to Predict
  Torque and Brake Specific Fuel Consumption of a Gasoline Engine,}
  \emph{Applied Energy}, Vol.~87, No.~11, 2010, pp. 3401--3408.
\newblock \doi{10.1016/j.apenergy.2010.04.027},
  \urlprefix\url{https://linkinghub.elsevier.com/retrieve/pii/S0306261910001340}.

\bibitem[{Bongard and Lipson(2007)}]{bongard2007automated}
Bongard, J., and Lipson, H., \enquote{Automated Reverse Engineering of
  Nonlinear Dynamical Systems,} \emph{Proceedings of the National Academy of
  Sciences}, Vol. 104, No.~24, 2007, pp. 9943--9948.
\newblock \doi{https://doi.org/10.1073/pnas.0609476104}.

\bibitem[{Schmelzer et~al.(2018)Schmelzer, Dwight, and
  Cinnella}]{schmelzer_data-driven_2018}
Schmelzer, M., Dwight, R., and Cinnella, P., \enquote{Data-{Driven}
  {Deterministic} {Symbolic} {Regression} of {Nonlinear} {Stress}-{Strain}
  {Relation} for {RANS} {Turbulence} {Modelling},} \emph{2018 {Fluid}
  {Dynamics} {Conference}}, {AIAA} {AVIATION} {Forum}, American Institute of
  Aeronautics and Astronautics, 2018, p.~13.
\newblock \doi{10.2514/6.2018-2900},
  \urlprefix\url{https://arc.aiaa.org/doi/10.2514/6.2018-2900}.

\bibitem[{Zhang and Smart(2004)}]{zhang2004multiclass}
Zhang, M., and Smart, W., \enquote{Multiclass Object Classification Using
  Genetic Programming,} \emph{Workshops on Applications of Evolutionary
  Computation}, Springer, Coimbra, Portugal, 2004, pp. 369--378.
\newblock \doi{https://doi.org/10.1007/978-3-540-31989-4_20}.

\bibitem[{Faris et~al.(2014)Faris, Al-Shboul, and Ghatasheh}]{faris2014genetic}
Faris, H., Al-Shboul, B., and Ghatasheh, N., \enquote{A Genetic Programming
  Based Framework for Churn Prediction in Telecommunication Industry,}
  \emph{International Conference on Computational Collective Intelligence},
  Springer, Seoul, Korea, 2014, pp. 353--362.
\newblock \doi{https://doi.org/10.1007/978-3-319-11289-3_36}.

\bibitem[{Faris et~al.(2013)Faris, Sheta, and
  {\"O}znergiz}]{faris2013modelling}
Faris, H., Sheta, A., and {\"O}znergiz, E., \enquote{Modelling Hot Rolling
  Manufacturing Process Using Soft Computing Techniques,} \emph{International
  Journal of Computer Integrated Manufacturing}, Vol.~26, No.~8, 2013, pp.
  762--771.
\newblock \doi{https://doi.org/10.1080/0951192X.2013.766937}.

\bibitem[{Forest et~al.(2018)Forest, Lacaille, Lebbah, and
  Azzag}]{forest2018generic}
Forest, F., Lacaille, J., Lebbah, M., and Azzag, H., \enquote{A Generic and
  Scalable Pipeline for Large-Scale Analytics of Continuous Aircraft Engine
  Data,} \emph{2018 IEEE International Conference on Big Data (Big Data)},
  IEEE, Seattle, WA, USA, 2018, pp. 1918--1924.
\newblock \doi{https://doi.org/10.1109/BigData.2018.8622297}.

\bibitem[{Searson(2015)}]{searson_gptips_2015}
Searson, D.~P., \enquote{{GPTIPS} 2: {An} {Open}-{Source} {Software} {Platform}
  for {Symbolic} {Data} {Mining},} \emph{Handbook of {Genetic} {Programming}
  {Applications}}, edited by A.~H. Gandomi, A.~H. Alavi, and C.~Ryan, Springer
  International Publishing, Cham, 2015, pp. 551--573.
\newblock \doi{10.1007/978-3-319-20883-1_22},
  \urlprefix\url{https://doi.org/10.1007/978-3-319-20883-1_22}.

\bibitem[{Shcherbakov et~al.(2013)Shcherbakov, Brebels, Shcherbakova, Tyukov,
  Janovsky, and Kamaev}]{shcherbakov2013survey}
Shcherbakov, M.~V., Brebels, A., Shcherbakova, N.~L., Tyukov, A.~P., Janovsky,
  T.~A., and Kamaev, V.~A., \enquote{A Survey of Forecast Error Measures,}
  \emph{World Applied Sciences Journal}, Vol.~24, No.~24, 2013, pp. 171--176.
\newblock \doi{10.5829/idosi.wasj.2013.24.itmies.80032}.

\end{thebibliography}

\end{document}

\maketitle

\begin{abstract}
Data-driven modeling is an imperative tool in various industrial applications, including many applications in the sectors of aeronautics and commercial aviation.
These models are in charge of providing key insights, such as which parameters are important on a specific measured outcome or which parameter values we should expect to observe given a set of input parameters. 
At the same time, however, these models rely heavily on assumptions (e.g., stationarity) or are ``black box'' (e.g., deep neural networks), meaning that they lack interpretability of their internal working and can be viewed only in terms of their inputs and outputs.
An interpretable alternative to the ``black box" models and with considerably less assumptions is symbolic regression (SR).
SR searches for the optimal model structure while simultaneously optimizing the model's parameters without relying on an a-priori model structure.\\
This document is supplemental material to the main paper ``An Application of Symbolic Regression on Exhaust Gas Temperature of Turbofan Engines: A Case Study'', by Kefalas, de Santiago Rojo Jr., Apostolidis, van den Herik, van Stein and B\"ack. 
It holds in detail all the results from the experiments performed together with plots and explanations.

\end{abstract}

\section{Introduction}\label{Introduction}

\lettrine{D}{ata}-driven modeling is an imperative tool in various industrial applications, including many applications in the sectors of aeronautics and commercial aviation. 
By data-driven modeling, we do not imply a conceptual model that is based on the data requirements of an application being developed, but rather a model of the underlying data-generating process. 
Such predictive models perform the task of identifying complex patterns in 
multimodal data, something that can also be loosely termed as reverse engineering~\cite{vaddireddy_feature_2020}.
These models are in charge of providing key insights, such as which parameters (covariates) are important on a specific measured outcome or which parameter values we should expect to observe given a set of input parameters. 
In addition, such models can infer future states of the system and distill new or refine existing physical models of nonlinear dynamical systems~\cite{vaddireddy_feature_2020}.

A physics-based model that adequately fits the data requires
a thorough understanding of the system’s physics and processes, which
can be prohibitively costly in terms of time and resources. 
On the other hand, there are cases where such models are necessary, especially in applications where no sufficient data have been generated yet. 
A good example is the design and certification phase of new aeronautical systems. 
Linear and non-linear statistical models rely on assumptions that might not hold (e.g., stationarity for ARMA ~\cite{box2015time} models in case of time-series). 
In contrast to those, non-parametric machine learning (ML) algorithms, such as the more recent deep neural networks (DNNs)~\cite{ian_goodfellow_deep_2016}
are considered to be ``black box'' models,
referring to processes  which  lack interpretability of their internal workings and can be viewed only in terms of their inputs and outputs. 
This means that these models do not explain their predictions/outputs in a way that is understandable by humans, and as a result, this lack of transparency and accountability can have severe consequences~\cite{rudin_stop_2019}, especially in safety-critical systems. 
However, model explainability is very important in a variety of engineering applications.

An \textit{interpretable} alternative to the ``black box" models and with considerably less assumptions is symbolic regression (SR). 
SR is a method for automatically finding a suitable algebraic expression that best describes the observed/sampled data \cite{koza1992genetic}. 
It is different from conventional regression techniques (e.g., linear regression, polynomial regression) in that SR does not rely on a specific a-priori model structure, but instead searches for the optimal model structure while simultaneously optimizing the model's parameters.  
The sole assumption made by SR is that the response surface \textit{can} be described algebraically~\cite{minnebo_empowering_nodate}. 
SR can be achieved by various methods, such as genetic programming (GP)~\cite{koza1992genetic,schmidt_distilling_nodate}, Bayesian methods~\cite{jin_bayesian_2020} and physics inspired artificial intelligence (AI)~\cite{udrescu_ai_2020}. 

Minimal human bias and low complexity of the modeling process
that allows function expressiveness and insights into the underlying data-generating process is of paramount importance. 
For aeronautical applications, safety is of the foremost significance and the consequences of failure or malfunction may be loss of life or serious injury, serious environmental damage, or harm to plant or property~\cite{lwears_rethinking_2012}.  
In aviation, properly understanding the data generating process can lead to developing  and improving existing physical models for non-linear dynamical systems that could lead to new insights, as well as indicate faults and failures that can save lives and money in the context of prognostics and health management (PHM)~\cite{nguyen_review_nodate}.

Nowadays, with the growing generation of large amounts of data in the aviation industry (e.g. passing from snapshot to continuous data collection), many applications have been developed and improved. 
Some of them\footnote{Predix Platform: \url{https://www.ge.com/digital/iiot-platform} \\ IntelligentEngine: \url{https://www.rolls-royce.com/products-and-services/civil-aerospace/intelligentengine.aspx}\\ The MRO Lab - Prognos: \url{https://www.afiklmem.com/en/solutions/about-prognos}}
are focused on engine health monitoring (EHM), as this is a central topic for engine manufacturers and operators.
Continuous engine operating data (CEOD) are collected at high frequencies in newer aircraft types, a development which -in combination with suitable algorithms- can improve the predictive capabilities for engine operators
With the purpose of improving the availability and operability of assets, EHM monitors the state of individual engines or engine fleets, by making use of historical operational data, or of data generated during past events.
By optimizing maintenance operations, not only safety is improved, but also asset utilization can be optimized, leading to reduced costs and improved operational efficiency. 
This is an area of interest not only for engine operators and maintenance providers, but also for engine manufacturers.
The aim of these data-driven solutions is primarily to avoid imminent failures by identifying possible anomalies in the engine operation, and secondly, to prevent over-maintenance of parts and components, exploiting their full life span. 

From an operational context, the use of models like the one presented in this paper can assist engine users to understand in depth the evolution of the deterioration of their engines, while making more reliable predictions about the time for maintenance actions and mitigate the possible disruptions in their flight and passenger operations. 
At the same, maintenance providers can predict the deterioration in detail and anticipate the physical state of the engines they will inspect and repair in the near future, without first having to wait for the real asset to be inducted in the shop. 
This way, they can streamline the maintenance process and provide more accurate quotations to their customers. 
Last, engine manufacturers -apart from being benefited in their maintenance business, for the aforementioned reasons-, they can use this type of work to understand in a better way the performance of their global fleet. 
This way, they can identify the influence of the different operating environments (e.g. presence of sand particles, salty water, air pollution, etc.) in the evolution of engines health and incorporate their findings in the design of either newer versions of the same engines or even to future engine generations.

The temperature of the exhaust gases of an engine, known as Exhaust Gas Temperature (EGT) has evolved to become the standard industrial indicator of the health of an aircraft engine \cite{von2014review}.  
This is because it can capture the cumulative effect of deterioration in the isentropic efficiency of gas path components.

The above information motivates our main research question: Can we uncover a \textit{meaningful} algebraic relationship between the EGT and the other measured parameters present in the CEOD 
data, using SR? By \textit{meaningful} we mean that the analytic expression between the EGT and the other parameters should also be justifiable.
The longtime industrial standard in engine health monitoring is the analysis of static, snapshot data. This approach, despite being computationally lighter, cannot capture the dynamics of continuous engine operation during the different flight phases
Having said that, our main contribution in this work is the first, to our knowledge, attempt to use real-world, \textit{continuous} data collected along the entire flight duration at a recording frequency of $1$Hz in order to model analytically the EGT against the rest of the monitored flight parameters.
These data, termed continuous engine operational data (CEOD), allow for a more complete digital representation of the operational history of an engine.

This document is supplemental material to the main paper ``An Application of Symbolic Regression on Exhaust Gas Temperature of Turbofan Engines: A Case Study''. 
It holds in detail all the results from the experiments performed together with plots and explanations.

The rest of this paper is organized as follows. In Sections~\ref{Appendix_A},\ref{Appendix_B},\ref{Appendix_C} we briefly describe the perfomed experiments and present in detail the experimental results, the generated formulas, and we show figures comparing the predicted to the ground truth EGT values of the test set.
Finally, in Section~\ref{sec:conclusion} we conclude this study.

\section{Experiment 1 Results}\label{Appendix_A}

For this experiment, a correlation analysis is performed upon the input variables, so that those which are highly correlated are discarded. The threshold used is $0.90$. Those above this threshold have been dropped, only keeping the first representative from a set of highly correlated variables. After these variables are deleted, $114$ remain to be used as final input variables in addition to the EGT.
The training, testing and validation error metrics results are displayed in tables \ref{tab:Exp1_train_error_metris}, \ref{tab:Exp1_test_error_metris}, and \ref{tab:Exp1_val_error_metris}  for all ten models resulting from the ten independent runs.

\begin{table}[h]
\centering
\resizebox{6cm}{!}{
\begin{tabular}{c|c|c|c}
\textbf{R\textasciicircum{}2} & \textbf{RMSE} & \textbf{MAE} & \textbf{MSE} \\ \hline
0.998102 & 1.239205 & 0.763866 & 1.535629 \\
0.997968 & 1.281986 & 0.786991 & 1.643489 \\
0.997507 & 1.420041 & 0.828125 & 2.016517 \\
0.997964 & 1.283411 & 0.786591 & 1.647145 \\
0.998020 & 1.265571 & 0.765595 & 1.601671 \\
0.998023 & 1.264679 & 0.738129 & 1.599412 \\
0.997745 & 1.350609 & 0.754343 & 1.824146 \\
0.998036 & 1.260616 & 0.746105 & 1.589153 \\
0.998123 & 1.232339 & 0.740111 & 1.518661 \\
0.997676 & 1.371248 & 0.791502 & 1.880321 \\ \hline
\end{tabular}}
\caption{Experiment 1 training error metrics.}
\label{tab:Exp1_train_error_metris}
\end{table}

\begin{table}[h]
\centering
\resizebox{6cm}{!}{
\begin{tabular}{c|c|c|c}
\textbf{R\textasciicircum{}2} & \textbf{RMSE} & \textbf{MAE} & \textbf{MSE} \\ \hline
0.99822 & 1.20770 & 0.75207 & 1.45853 \\
0.99803 & 1.27203 & 0.77908 & 1.61805 \\
0.99767 & 1.38459 & 0.82148 & 1.91708 \\
0.99808 & 1.25604 & 0.77368 & 1.57764 \\
0.99810 & 1.24747 & 0.75516 & 1.55618 \\
0.99813 & 1.24030 & 0.72427 & 1.53834 \\
0.99788 & 1.32030 & 0.74576 & 1.74320 \\
0.99817 & 1.22514 & 0.73863 & 1.50097 \\
0.99814 & 1.23560 & 0.73759 & 1.52672 \\
0.99783 & 1.33519 & 0.78411 & 1.78273 \\ \hline
\end{tabular}}
\caption{Experiment 1 test error metrics.}
\label{tab:Exp1_test_error_metris}
\end{table}

\begin{table}[h!]
\centering
\resizebox{6.5cm}{!}{
\begin{tabular}{c|c|c|c}
\textbf{R\textasciicircum{}2} & \textbf{RMSE} & \textbf{MAE} & \textbf{MSE} \\ \hline
0.981343 & 3.240743 & 1.357668 & 10.502417  \\
0.947178 & 5.452943  & 2.940967 & 29.734585   \\
0.808921 & 10.371179 & 3.410746 & 107.561345 \\
0.984409 & 2.962510  & 1.224247 & 8.776463 \\
0.754388 & 11.758352 & 3.642960 & 138.258840 \\
0.805126 & 10.473665 & 3.333293 & 109.697659 \\
0.812055 & 10.285767 & 3.238445 & 105.797008 \\
0.809934 & 10.343668 & 3.325208 & 106.991471 \\
0.855140 & 9.030162  & 4.563011 & 81.543823 \\
0.805914 & 10.452461 & 3.109154 & 109.253937 \\ \hline
\end{tabular}}
\caption{Experiment 1 validation error metrics.}
\label{tab:Exp1_val_error_metris}
\end{table}

\newpage
Each generated symbolic expression model is shown below:
\\

$Y_{1}^{1} = 0.141\cdot x_{4} + 0.123\cdot x_{5} + 0.8\cdot x_{6} + 0.0214\cdot x_{12} - 0.123\cdot x_{18} + 0.751\cdot x_{21} + 0.0261x_{60} + 0.0405\cdot x_{74} - 0.0371\cdot |log(x_{39})| + 1.32\cdot10^{-4} e^{2\cdot x_{21}} - 0.0428\cdot |x_{4}| - 0.0261\cdot e^{x_{21}} + 0.00762\cdot {x_{74}}^2 + 0.133$
\\

$Y_{1}^{2} = 0.13\cdot x_{4} + 0.668\cdot x_{6} + 0.0264\cdot x_{12} - 0.0796\cdot x_{14} + 0.759\cdot x_{21} + 0.0484\cdot x_{43} - 0.00864\cdot x_{57} + 0.0219\cdot x_{110} - 0.0219\cdot x_{113}  - 0.0264\cdot |x_{4}| 
+ 0.0219\cdot |x_{6}| - 0.00702\cdot |x_{43} - 2.0\cdot x_{46} + 0.2\cdot {x_{21}}^3| - 0.00579\cdot {x_{21}}^3 - 6.3\cdot10^{-4}\cdot|x_{14} - 
1.0\cdot x_{21}|\cdot|x_{21}|^3 \cdot|x_{43}| + 0.0119$
\\

$Y_{1}^{3} = 0.169\cdot x_{4} - 0.0168\cdot x_{1} + 0.125\cdot x_{5} + 0.712\cdot x_{6} - 0.0336\cdot x_{18} + 0.704\cdot x_{21} + 0.0376\cdot x_{43} + 0.0156\cdot x_{110} - 0.00778\cdot e^{-x_{12}} - 0.0808\cdot |x_{4}| - 0.0513\cdot |x_{8}| - 0.0156\cdot |x_{21}| + 0.0432\cdot {|x_{4}|}^{1/2} - 0.00102\cdot |x_{21} + |(x_{21})||^3 + 0.0368 (|x_{74}|^3)^{1/2} + 0.0736$
\\

$Y_{1}^{4} = 0.0758\cdot x_{4} + 0.885\cdot x_{6} - 0.2\cdot x_{13} - 0.156\cdot x_{14} + 0.43\cdot x_{21} - 0.0244\cdot x_{23} - 0.495\cdot e^{(-e^{(e^{(x_{43})})})} + 0.2\cdot e^{(-e^{(-x_{4})})}+ 0.156\cdot e^{(-e^{(-x_{14})})} + 0.00361\cdot e^{(x_{110})} - 0.787\cdot e^{(-e^{(x_{21})})} + 0.00719\cdot {x_{14}}^2 + 0.214$
\\

$Y_{1}^{5} = 0.0887\cdot x_{4} + 0.111\cdot x_{5} + 0.69\cdot x_{6} + 0.0296\cdot x_{11} + 0.523\cdot x_{21} - 0.0442\cdot x_{39} - 0.121\cdot e^{(- x_{39})} - 0.499\cdot e^{(-e^{(x_{21})})} + 0.0559\cdot x_{4} {({e^{(x_{5})}})}^{1/2} - 0.00493\cdot x_{74}\cdot x_{113} + 0.353$
\\

$Y_{1}^{6} = 0.204\cdot x_{4} + 0.592\cdot x_{6} + 0.0242\cdot x_{11} + 0.485\cdot x_{21} + 0.0698\cdot x_{43} + 0.122\cdot x_{59} - 0.642\cdot e^{(-e^{(x_{21})})} - (0.00133\cdot|x_{43}|)/|x_{94}|^2 - 0.0205\cdot {x_{4}}^2 + 0.00106\cdot {x_{4}}^3 + 0.365$
\\

$Y_{1}^{7} = 0.111\cdot x_{4} + 0.601\cdot x_{6} + 0.0264\cdot x_{11} + 0.448\cdot x_{21} - 0.0361\cdot x_{25} + 0.0459\cdot x_{43} + 0.111\cdot x_{59} - 0.0346\cdot x_{113} - 0.0553\cdot e^{(-x_{4})} + 0.02\cdot e^{(x_{113})} - 0.774\cdot e^{(-e^{(x_{21})})} + 0.326$
\\

$Y_{1}^{8} = 0.133\cdot x_{4} + 0.649\cdot x_{6} + 0.00805\cdot x_{12} - 0.101\cdot x_{14} + 0.749\cdot x_{21} - 0.0178\cdot x_{23} + 0.0268\cdot x_{24} + 0.0463\cdot x_{43} + 0.00805\cdot x_{52} + 0.0155\cdot x_{74} + 0.0268\cdot x_{110} - 0.0115\cdot x_{111} - 0.0178\cdot x_{113} - 0.0178\cdot |x_{4}| - 0.0178\cdot |x_{74}| + 0.00455\cdot x_{4}\cdot x_{6} - 0.0176\cdot x_{21}\cdot x_{74} + 0.0176\cdot x_{23}\cdot x_{74} - 0.0176\cdot x_{52}\cdot x_{74} + 0.0176\cdot x_{74} |x_{21}| - 0.00228\cdot {x_{4}}^2 - 0.00228\cdot {x_{6}}^2 - 0.0311\cdot {x_{21}}^2 + 0.0271$
\\

$Y_{1}^{9} = 0.136\cdot x_{4} + 0.592\cdot x_{6} + 0.701\cdot x_{21} + 0.0567\cdot x_{43} + 0.127\cdot x_{59} + 0.0291\cdot x_{96} - 0.0263\cdot x_{113} - 0.00304|(x_{8} - 5.11)|\cdot |x_{21}|^2 + 0.00248\cdot x_{4} x_{59} - 0.0307\cdot x_{21}\cdot |x_{21}| - 4.08\cdot10^{-4}\cdot {x_{113}}^3 + 0.0092$
\\

$Y_{1}^{10} = 0.17\cdot x_{4} - 0.00532\cdot x_{3} + 0.121\cdot x_{5} + 0.685\cdot x_{6} + 0.0105\cdot x_{12} + 0.65\cdot x_{21} + 0.0375\cdot x_{24} - 0.00532\cdot x_{39} + 0.075\cdot x_{74} - 0.0079\cdot x_{112} - 0.00532\cdot x_{43}/x_{94} - 0.0108\cdot {x_{4}}^2 + 4.7\cdot10^{-4}\cdot {x_{4}}^3 - 0.0202\cdot {x_{21}}^2 - 0.00258\cdot {x_{21}}^3 - 0.00258\cdot {x_{39}}^2 + 0.0246$

\newpage
In the figures below we show the model predictions against the actual values of the EGT per model on the test set:

\begin{figure}[!h]
\centering
\subfloat[Model 1]{\includegraphics[width = 3in]{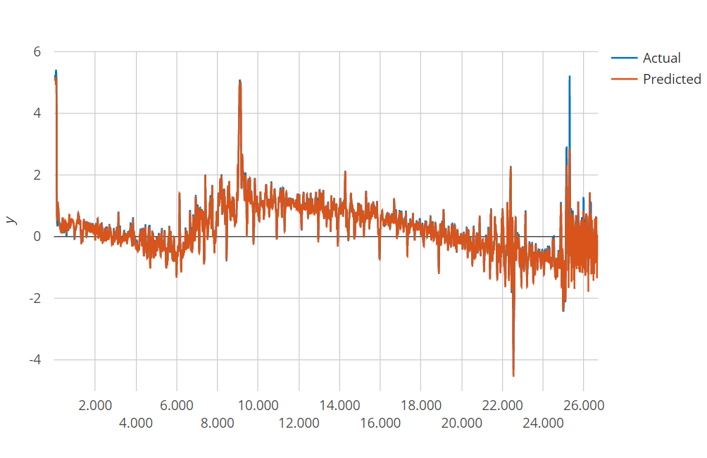}}
\subfloat[Model 2]{\includegraphics[width = 3in]{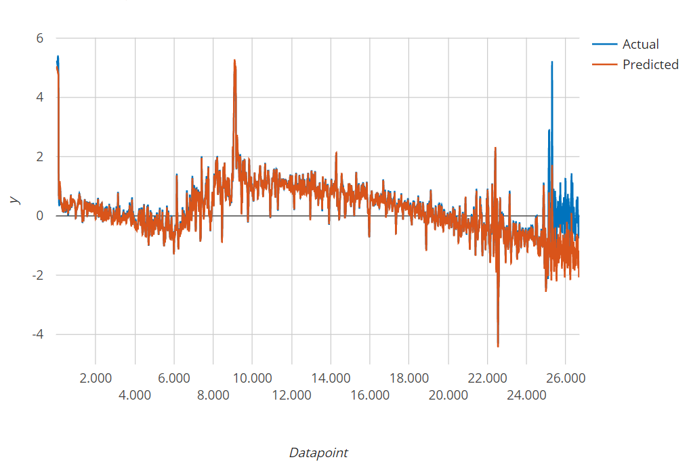}}\\
\subfloat[Model 3]{\includegraphics[width = 3in]{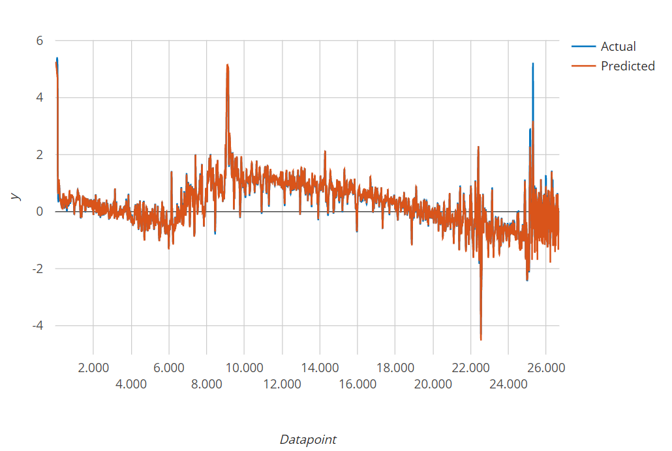}}
\subfloat[Model 4]{\includegraphics[width = 3in]{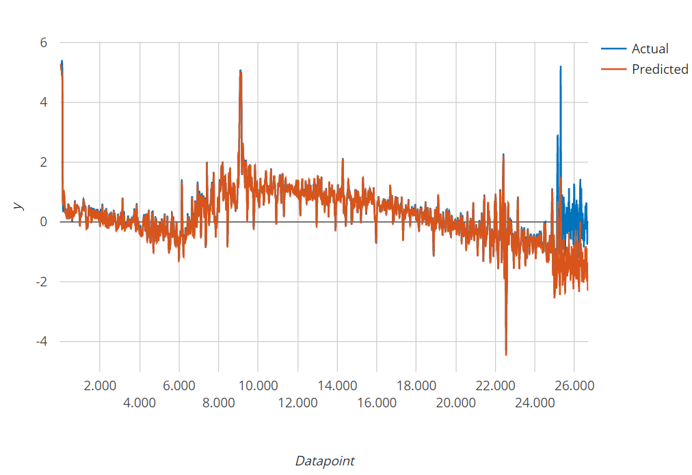}}
\caption{Scaled EGT predictions (red) ($y$-axis) on the validation set 
    vs.~observed (blue) values (Experiment 1). $x$-axis shows the data sample index in consecutive order according to their sampling over time.}
\label{ModelsExp1.1-4}
\end{figure}

\begin{figure}[!h]
\centering
\subfloat[Model 5]{\includegraphics[width = 3in]{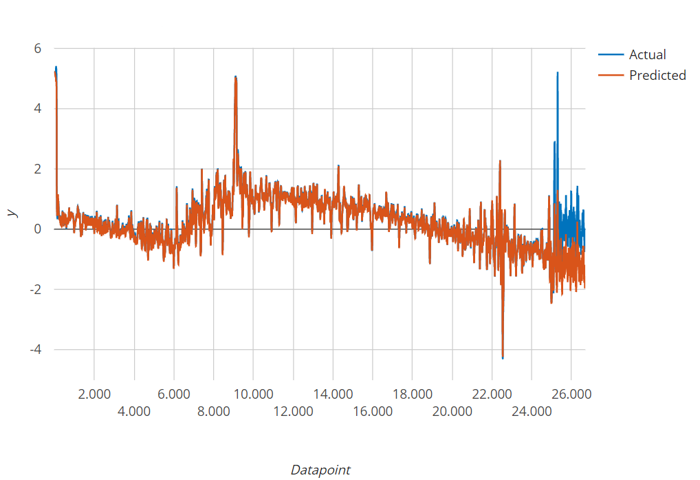}}
\subfloat[Model 6]{\includegraphics[width = 3in]{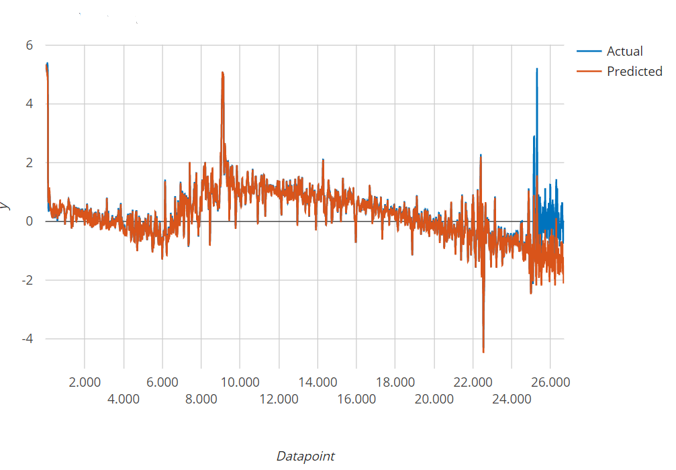}}\\
\subfloat[Model 7]{\includegraphics[width = 3in]{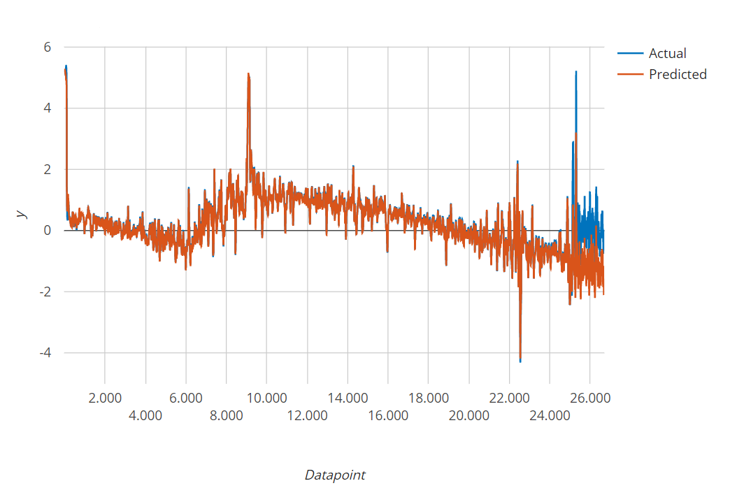}}
\subfloat[Model 8]{\includegraphics[width = 3in]{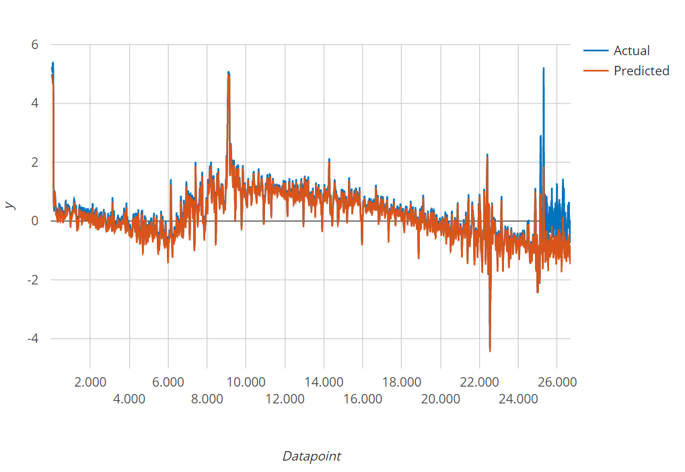}}\\
\subfloat[Model 9]{\includegraphics[width = 3in]{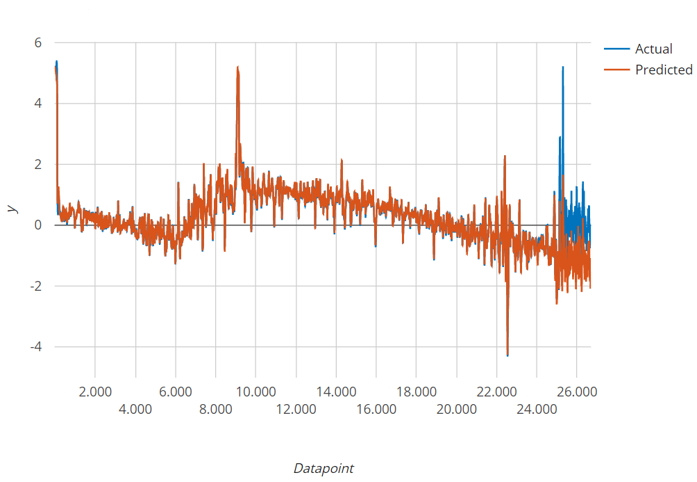}}
\subfloat[Model 10]{\includegraphics[width = 3in]{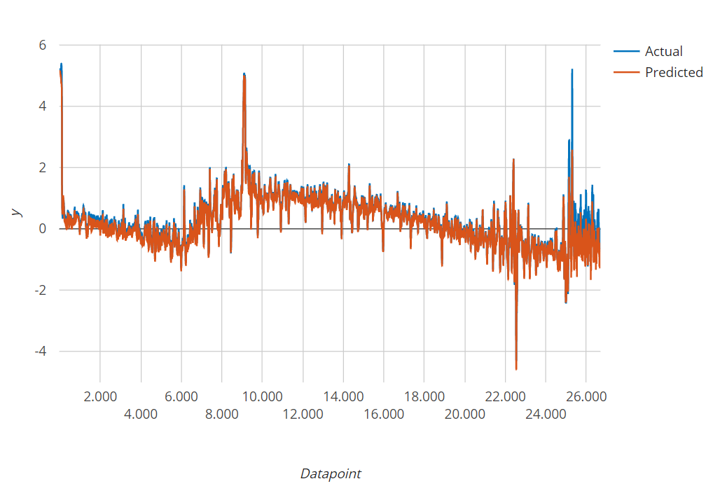}}
\caption{Scaled EGT predictions (red) ($y$-axis) on the validation set 
    vs.~observed (blue) values (Experiment 1). $x$-axis shows the data sample index in consecutive order according to their sampling over time.}
\label{ModelsExp1.5-8}
\end{figure}

\newpage
\section{Experiment 2 Results}\label{Appendix_B}

Once experiment $1$ was completed and results were analyzed, the need for developing a control model is satisfied by experiment $2$. Here, one highly correlated variable, which is not above the proposed threshold, was dropped from the input variables. This variables is the \textit{Average Temperature at Station 25 (DEG\_C)}. Therefore, $113$ input variables were used in GPTIPS in order to model the EGT.
The training and validation error metrics results are displayed in tables \ref{tab:Exp2_train_error_metris}, \ref{tab:Exp2_test_error_metris}, \ref{tab:Exp2_val_error_metris} and  for all ten models resulting from the ten independent runs.

\begin{table}[h]
\centering
\resizebox{6cm}{!}{
\begin{tabular}{c|c|c|c}
\textbf{R\textasciicircum{}2} & \textbf{RMSE} & \textbf{MAE} & \textbf{MSE} \\ \hline
0.99768 & 1.36893 & 0.85751 & 1.87396 \\
0.99756 & 1.40389 & 0.88441 & 1.97090 \\
0.99751 & 1.41842 & 0.87606 & 2.01192 \\
0.99696 & 1.56913 & 0.95831 & 2.46217 \\
0.99743 & 1.44187 & 0.87915 & 2.07899 \\
0.99765 & 1.37774 & 0.85544 & 1.89818 \\
0.99741 & 1.44669 & 0.86240 & 2.09291 \\
0.99756 & 1.40450 & 0.87028 & 1.97263 \\
0.99746 & 1.43365 & 0.86886 & 2.05535 \\
0.99762 & 1.38674 & 0.85581 & 1.92304 \\ \hline
\end{tabular}}
\caption{Experiment 2 training error metrics.}
\label{tab:Exp2_train_error_metris}
\end{table}

\begin{table}[h]
\centering
\resizebox{6cm}{!}{
\begin{tabular}{c|c|c|c}
\textbf{R\textasciicircum{}2} & \textbf{RMSE} & \textbf{MAE} & \textbf{MSE} \\ \hline
0.99781 & 1.34160 & 0.84267 & 1.79988 \\
0.99766 & 1.38506 & 0.87861 & 1.91840 \\
0.99766 & 1.38600 & 0.86851 & 1.92098 \\
0.99701 & 1.56784 & 0.95604 & 2.45812 \\
0.99751 & 1.42995 & 0.87120 & 2.04477 \\
0.99784 & 1.33301 & 0.84164 & 1.77691 \\
0.99748 & 1.43730 & 0.85617 & 2.06582 \\
0.99767 & 1.38220 & 0.86608 & 1.91047 \\
0.99755 & 1.41698 & 0.86252 & 2.00783 \\
0.99777 & 1.35305 & 0.84415 & 1.83074 \\ \hline
\end{tabular}}
\caption{Experiment 2 test error metrics.}
\label{tab:Exp2_test_error_metris}
\end{table}

\begin{table}[h!]
\centering
\resizebox{6.5cm}{!}{
\begin{tabular}{c|c|c|c}
\textbf{R\textasciicircum{}2} & \textbf{RMSE} & \textbf{MAE} & \textbf{MSE} \\ \hline
0.760114 & 11.62049 & 3.508928 & 135.0357 \\
0.796653 & 10.69894 & 3.228156 & 114.4673 \\
0.781762 & 11.08375 & 3.578363 & 122.8495 \\
0.879283 & 8.243387 & 3.037525 & 67.95342 \\
0.805173 & 10.4724  & 3.198549 & 109.6712 \\
0.794995 & 10.74246 & 3.204305 & 115.4005 \\
0.78722  & 10.94429 & 3.307758 & 119.7775 \\
0.809058 & 10.36747 & 3.247515 & 107.4844 \\
0.79427  & 10.76145 & 3.539824 & 115.8089 \\
0.891921 & 7.799965 & 2.70033  & 60.83946 \\ \hline
\end{tabular}}
\caption{Experiment 2 validation error metrics.}
\label{tab:Exp2_val_error_metris}
\end{table}

\newpage
Each generated symbolic expression model is shown below:
\\

$Y_{2}^{1} = 0.0789\cdot x_{4} + 0.143\cdot x_{5} + 0.0207\cdot x_{10} + 0.799\cdot x_{17} - 0.524\cdot x_{19} + 0.575\cdot x_{20} - 0.0789\cdot e^{(-e^{(- x_{20})})} - 0.441\cdot e^{(-e^{(-e^{(-x_{4})})})} - e^{(-e^{(-e^{(-x_{20})})})} - ((0.0186\cdot |x_{38}|)/|x_{98}|) + ((0.0092\cdot x_{38})/x_{51}) + 1.13$
\\ 

$Y_{2}^{2} = 0.13\cdot x_{4} + 0.0229\cdot x_{11} + 0.694\cdot x_{17} - 0.394\cdot x_{19} + 0.723\cdot x_{20} + 0.0604\cdot x_{42} + 0.107\cdot x_{58} - 0.0261\cdot |x_{19}|\cdot |x_{20}| + 0.0168\cdot x_{42}\cdot x_{112} - 0.033\cdot x_{20}\cdot |x_{20}|- 0.0162\cdot x_{42}\cdot |x_{20}| - 0.013\cdot x_{112}\cdot |x_{19}| + 0.0126$
\\

$Y_{2}^{3} = 0.123\cdot x_{4} + 0.0305\cdot x_{10} + 0.713\cdot x_{17} - 0.327\cdot x_{19} + 0.628\cdot x_{20} + 0.0582\cdot x_{42} + 0.0949\cdot x_{58} + 0.0263\cdot x_{73} - 0.0303\cdot x_{112} - 0.0119\cdot {x_{20}}^2\cdot x_{42} + 0.00167$
\\ 

$Y_{2}^{4} = 0.1\cdot x_{4} + 0.671\cdot x_{17} - 0.389\cdot x_{19} + 0.506\cdot x_{20} - 0.0408\cdot x_{38} + 0.138\cdot x_{58} - 0.108\cdot e^{(-x_{38})} + 0.929\cdot e^{(-e^{(-e^{(-e^{(-x_{4})})})})} - 0.0472\cdot |x_{7}| - 0.633\cdot e^{(-e^{(x_{20})})} - 0.0392$
\\ 

$Y_{2}^{5} = 0.11\cdot x_{4} + 0.662\cdot x_{17} - 0.39\cdot x_{19} + 0.526\cdot x_{20} + 0.0537\cdot x_{42} + 0.141\cdot x_{58} + 0.0269\cdot x_{95} - 0.0609\cdot e^{(-x_{4})} - 0.00223\cdot e^{(x_{20})} - 0.587\cdot e^{(-1.1e^{(x_{20})})} - 0.00143\cdot x_{23}\cdot {x_{112}}^2 - 0.00143\cdot {x_{23}}^2\cdot x_{112} - 4.76\cdot10^4\cdot {x_{23}}^3 - 4.76\cdot10^4\cdot {x_{112}}^3 + 0.276$
\\ 

$Y_{2}^{6} = 0.647\cdot x_{17} - 0.412\cdot x_{19} + 0.549\cdot x_{20} + 0.0505\cdot x_{42} + 0.157\cdot x_{58} + 0.0243\cdot x_{95} - 0.826\cdot e^{(-e^{(-e^{(-x_{20})})})} - 1.59\cdot e^{(-e^{(- e^{{(-x_{4})}^{1/2}})})} + 0.00578\cdot {x_{4}}^2 - 7.58\cdot10^5\cdot {x_{51}}^4 + 1.69$
\\ 

$Y_{2}^{7} = 0.0925\cdot x_{4} + 0.113\cdot x_{5} + 0.0195\cdot x_{11} + 0.797\cdot x_{17} - 0.427\cdot x_{19} + 0.518\cdot x_{20} - 0.0181\cdot x_{112} - 0.452\cdot e^{(-x_{38})} - 1.13\cdot e^{(-e^{(- x_{38})})} - 0.45\cdot e^{(-e^{(-e^{(-x_{4})})})} - 0.847\cdot e^{(-e^{(-e^{(-x_{20})})})} + 1.89$
\\ 

$Y_{2}^{8} = 0.158\cdot x_{4} + 0.131\cdot x_{5} + 0.8\cdot x_{17} - 0.46\cdot x_{19} + 0.53\cdot x_{20} - 0.0231\cdot x_{22} - 0.0231\cdot e^{(-x_{4})} - 0.0826\cdot e^{(-x_{38})} - 0.0826\cdot e^{(-2\cdot e^{(- x_{20})})} - 0.928\cdot e^{(-e^{(-e^{(-x_{20})})})} - 0.00491\cdot {x_{4}}^2 - 0.00827\cdot {x_{38}}^2 + 0.811$
\\ 

$Y_{2}^{9} = 0.172\cdot x_{4} + 0.0239\cdot x_{10} + 0.667\cdot x_{17} - 0.403\cdot x_{19} + 0.593\cdot x_{20} + 0.0475\cdot x_{42} + 0.133\cdot x_{58} + 0.0481\cdot x_{73} - 0.0481\cdot e^{(0.567\cdot x_{20})} - 0.623\cdot e^{(- e^{0.73\cdot x_{20}})} - 0.0647\cdot |x_{4}| + 0.316$
\\ 

$Y_{2}^{10} = 0.129\cdot x_{4} + 0.69\cdot x_{17} - 0.373\cdot x_{19} + 0.455\cdot x_{20} - 0.0382\cdot x_{24} + 0.121\cdot x_{58} - 0.0303\cdot x_{112} + 0.509\cdot e^{(- e^{(- e^{(- e^{(- x_{42})})})})} + 0.0771\cdot e^{(-2\cdot {x_{7}}^2)} - 0.707\cdot e^{(- e^{x_{20}})} - 0.0137$

\newpage
In the figures below we show the model predictions against the actual values of the EGT per model on the test set:

\begin{figure}[h]
\centering
\subfloat[Model 1]{\includegraphics[width = 3in]{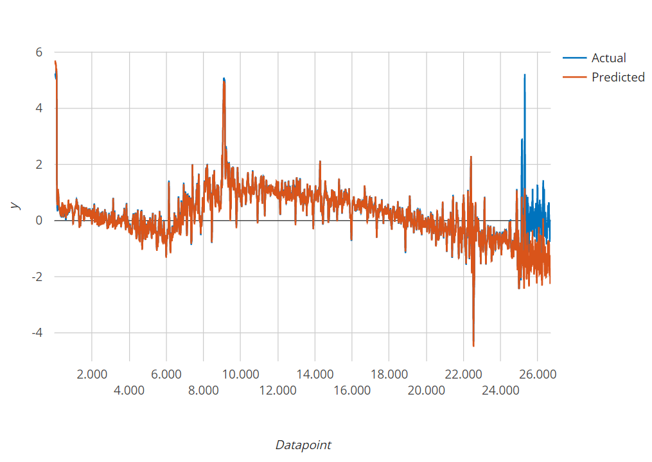}} 
\subfloat[Model 2]{\includegraphics[width = 3in]{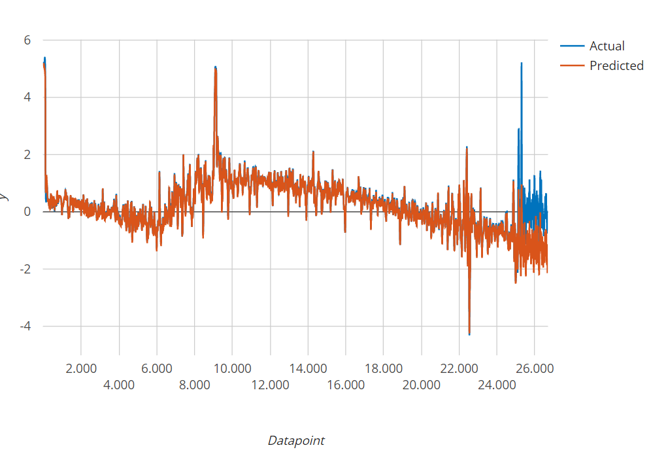}}\\ 
\subfloat[Model 3]{\includegraphics[width = 3in]{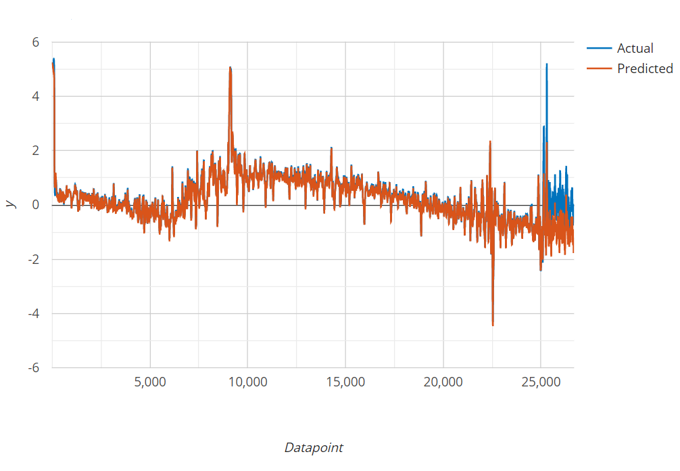}}
\subfloat[Model 4]{\includegraphics[width = 3in]{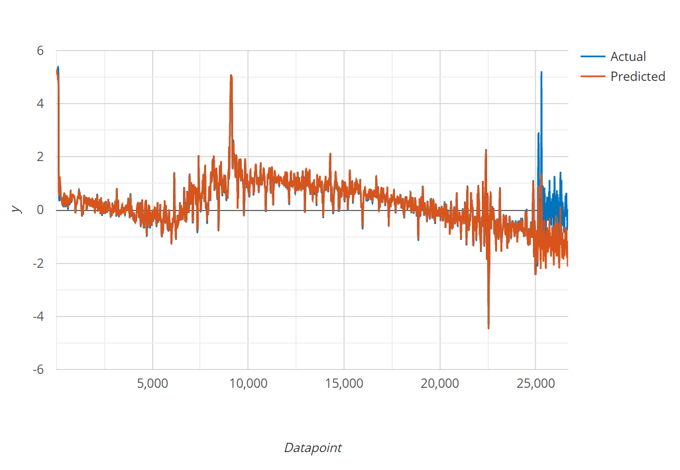}}\\

\caption{Scaled EGT predictions (red) ($y$-axis) on the validation set 
    vs.~observed (blue) values (Experiment 2). $x$-axis shows the data sample index in consecutive order according to their sampling over time.}
\label{ModelsExp2.2-7}
\end{figure}

\begin{figure}[!h]
\centering
\subfloat[Model 5]{\includegraphics[width = 3in]{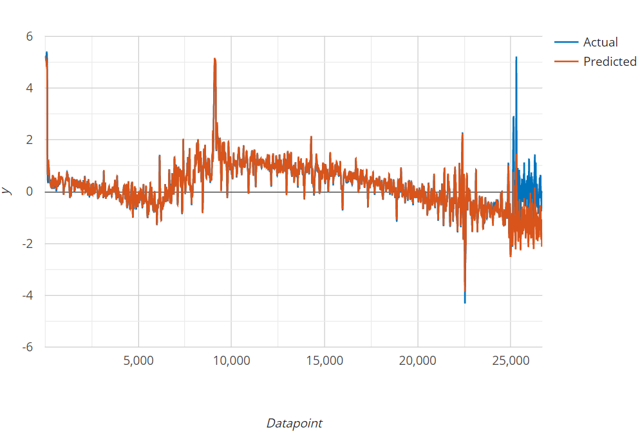}}
\subfloat[Model 6]{\includegraphics[width = 3in]{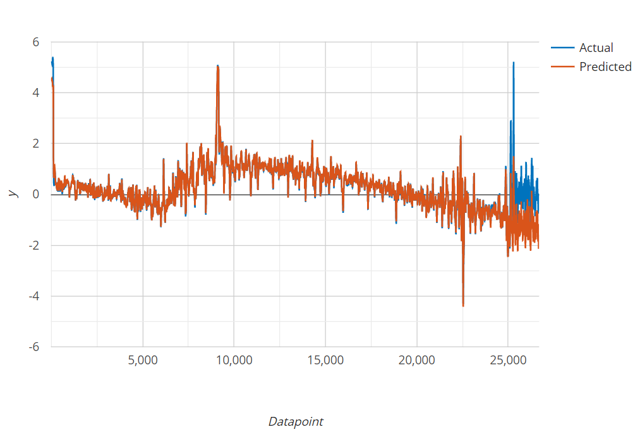}}\\
\subfloat[Model 7]{\includegraphics[width = 3in]{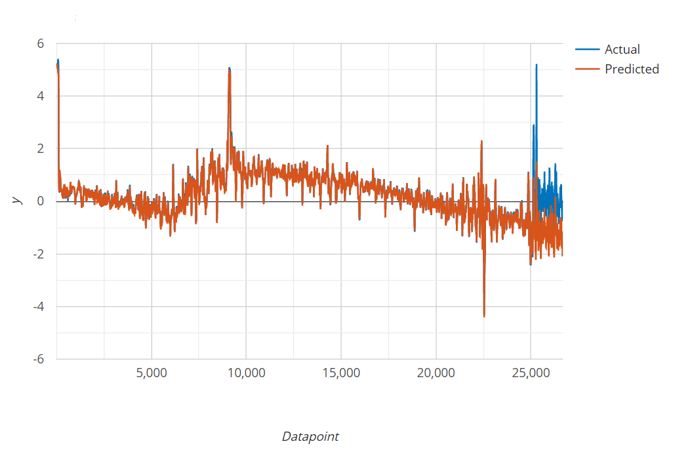}}
\subfloat[Model 8]{\includegraphics[width = 3in]{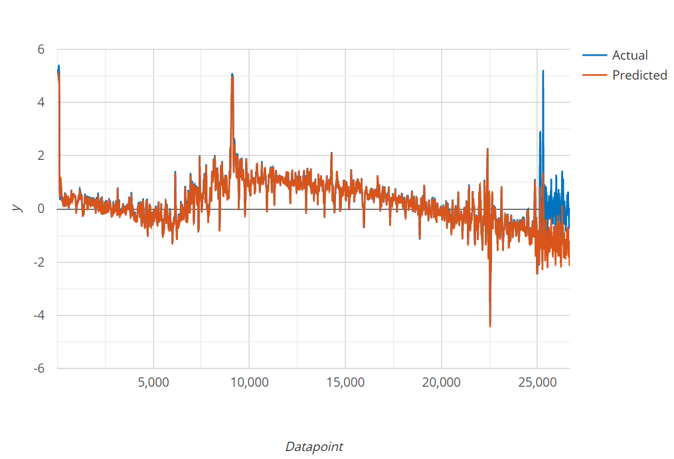}}\\
\subfloat[Model 9]{\includegraphics[width = 3in]{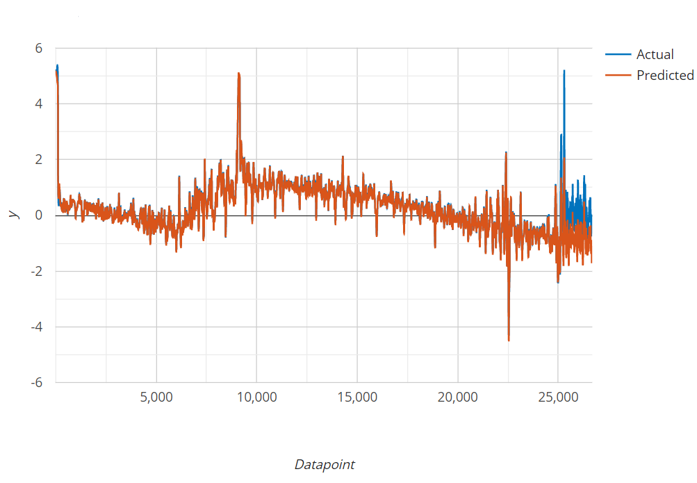}}
\subfloat[Model 10]{\includegraphics[width = 3in]{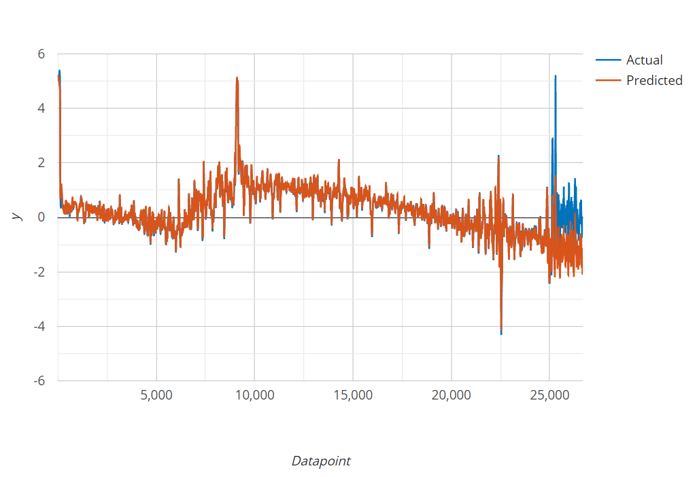}}
\caption{Scaled EGT predictions (red) ($y$-axis) on the validation set 
    vs.~observed (blue) values (Experiment 2). $x$-axis shows the data sample index in consecutive order according to their sampling over time.}
\label{ModelsExp2.8-10}
\end{figure}

\newpage

\section{Experiment 3 Results}\label{Appendix_C}

A correlation analysis between the input variables and the output variable, EGT, is performed, dropping those variables whose correlation with EGT is higher than $0.90$. After doing so, $186$ input variables remained. Once this was done, and following experiment's $1$ logic, a correlation analysis upon the remaining input variables is done. Again, those above the proposed threshold are deleted, only keeping the first representative from a set of highly correlated variables. Once this step was done, $112$ input variables remained, plus the output variable, EGT.
The training and validation error metrics results are displayed in tables \ref{tab:Exp3_training_error_metris}, \ref{tab:Exp3_test_error_metris}, \ref{tab:Exp3_val_error_metris} and  for all ten models resulting from the ten independent runs.

\begin{table}[h]
\centering
\resizebox{6cm}{!}{
\begin{tabular}{c|c|c|c}
\textbf{R\textasciicircum{}2} & \textbf{RMSE} & \textbf{MAE} & \textbf{MSE} \\ \hline
0.99800 & 1.27580 & 0.74480 & 1.62768 \\
0.99812 & 1.23542 & 0.75659 & 1.52627 \\
0.99807 & 1.25291 & 0.76230 & 1.56978 \\
0.99822 & 1.20218 & 0.75237 & 1.44524 \\
0.99809 & 1.24816 & 0.73935 & 1.55789 \\
0.99832 & 1.16893 & 0.75271 & 1.36640 \\
0.99801 & 1.27283 & 0.78877 & 1.62009 \\
0.99818 & 1.21660 & 0.73428 & 1.48011 \\
0.99820 & 1.21099 & 0.72872 & 1.46649 \\
0.99835 & 1.16035 & 0.71518 & 1.34641 \\ \hline
\end{tabular}}
\caption{Experiment 3 training error metrics.}
\label{tab:Exp3_training_error_metris}
\end{table}

\begin{table}[h]
\centering
\resizebox{6cm}{!}{
\begin{tabular}{c|c|c|c}
\textbf{R\textasciicircum{}2} & \textbf{RMSE} & \textbf{MAE} & \textbf{MSE} \\ \hline
0.99792  & 1.29194  & 0.74879  & 1.66910  \\
0.99799  & 1.26866  & 0.76180  & 1.60950  \\
0.99801  & 1.26294  & 0.75916  & 1.59503  \\
0.99810  & 1.23463  & 0.75536  & 1.52432  \\
0.99802  & 1.26004  & 0.74229  & 1.58769  \\
0.99823  & 1.19080  & 0.75385  & 1.41801  \\
0.99793  & 1.28963  & 0.78988  & 1.66315  \\
0.99812  & 1.22731  & 0.73710  & 1.50629  \\
0.99813  & 1.22366  & 0.73682  & 1.49735  \\
0.998197 & 1.202644 & 0.720882 & 1.446352 \\ \hline
\end{tabular}}
\caption{Experiment 3 test error metrics.}
\label{tab:Exp3_test_error_metris}
\end{table}

\begin{table}[h!]
\centering
\resizebox{6.5cm}{!}{
\begin{tabular}{c|c|c|c}
\textbf{R\textasciicircum{}2} & \textbf{RMSE} & \textbf{MAE} & \textbf{MSE} \\ \hline
0.788536 & 10.91038 & 3.23724 & 119.03647 \\
0.834811 & 9.64300  & 2.97148 & 92.98735  \\
0.985225 & 2.88393  & 1.18922 & 8.31704   \\
0.985034 & 2.90254  & 1.31293 & 8.42474   \\
0.800509 & 10.59703 & 3.35942 & 112.29699 \\
0.819613 & 10.07684 & 3.18990 & 101.54277 \\
0.802258 & 10.55046 & 3.35021 & 111.31211 \\
0.933994 & 6.09554  & 2.08652 & 37.15565  \\
0.794766 & 10.74848 & 4.10824 & 115.52974 \\
0.822608 & 9.99285  & 3.21731 & 99.85703 \\ \hline
\end{tabular}}
\caption{Experiment 3 validation error metrics.}
\label{tab:Exp3_val_error_metris}
\end{table}

\newpage
Each generated symbolic expression model is shown below:

$Y_{3}^{1} = 0.114\cdot x_{5} + 0.771\cdot x_{6} + 0.0249\cdot x_{11} - 0.0763\cdot x13 + 0.395\cdot x_{19} + 1.62\cdot log(x_{4} + e^{(-x_{23})} + 8.9) + 0.233\cdot e^{(-x_{23})} - 0.0534\cdot e^{(-x_{41})} - 1.36\cdot{10}^{-5} e^{(-x_{103})} - 1.14\cdot e^{(-e^{(-e^{(-x_{19})})})} - 3.11$
\\
 
$Y_{3}^{2} = 0.0789\cdot x_{4} + 0.123\cdot x_{5} + 0.696\cdot x_{6} + 0.0219\cdot x_{11} + 0.68\cdot x_{19} + 0.156\cdot x_{72} + 0.357\cdot e^{(-e^{(-x_{4})})} + 0.29\cdot e^{(-e^{(-x_{37})})} + 0.54\cdot e^{({-x_{19}}^{4} \cdot {x_{72}}^{2})} + 0.393\cdot e^{(-{x_{37}}^{2})} - 0.996$
\\

$Y_{3}^{3} = 0.0765\cdot x_{4} + 0.683\cdot x_{6} - 0.114\cdot x_{14} + 0.683\cdot x_{19} - 0.0189\cdot x_{21} - 0.0228\cdot x_{40} + 0.0371\cdot x_{41} + 0.159\cdot x_{72} - 0.585\cdot e^{(-e^{(-e^{(-x_{4})})})} + 0.0371\cdot e^{(-e^{(-e^{(-x_{26})})})} - 0.136\cdot |x_{19} - 1.23| + 0.114\cdot e^{(-real(e^{(-x_{26})}))} - 0.699\cdot e^{(-real(e^{(-x_{72})}))} + 0.766$
\\

$Y_{3}^{4} = 0.139\cdot x_{4} + 0.107\cdot x_{5} + 0.698\cdot x_{6} + 0.438\cdot x_{19} - 0.039\cdot x_{23} + 0.0508\cdot x_{41} - 0.0265\cdot x_{111} - 0.145\cdot e^{(-e^{(-e^{(-x_{11})})})} - 0.825\cdot e^{(-e^{(x_{19})})} - 2.66\cdot10{-4}\cdot {x_{4}}^{3} + 0.411$
\\

$Y_{3}^{5} = 0.0906\cdot x_{4} + 0.642\cdot x_{6} + 0.028\cdot x_{11} + 0.562\cdot x_{19} + 0.0752\cdot x_{37} + 0.0171\cdot x_{67} + 0.0338\cdot x_{74} - 0.791\cdot e^{(-Re(e^{(-e^{(-x_{19})})}))} - 0.115\cdot |x_{37}| + (0.208\cdot{0.657}^{x_{14}} \cdot{0.657}^{x_{74}})/0.657^{x_{108}} + 0.417$
\\

$Y_{3}^{6} = 0.101\cdot x_{4} + 0.136\cdot x_{5} + 0.654\cdot x_{6} + 0.448\cdot x_{19} - 0.0226\cdot x_{21} + 0.0426\cdot x_{26} + 0.0484\cdot x_{41} - 0.0602\cdot x_{74} + 0.082\cdot x_{108} - 0.605\cdot e^{(-e^{(x_{19})})} + 0.224
$
\\

$Y_{3}^{7} = 0.0831\cdot x_{4} + 0.117\cdot x_{5} + 0.695\cdot x_{6} + 0.0233\cdot x_{11} + 0.461\cdot x_{19} - 0.0436\cdot x_{23} + 0.0436\cdot x_{41} + 0.0544\cdot x_{72} - 0.729\cdot e^{(e^{(-e^{(x_{19})})})} + 0.81\cdot e^{(-e^{(e^{(x_{19})})})} + 0.368\cdot e^{(-e^{(-x_{4})})} + 0.883
$
\\

$Y_{3}^{8} = 0.0596\cdot x_{4} + 0.118\cdot x_{5} + 0.671\cdot x_{6} + 0.474\cdot x_{19} + 0.0446\cdot x_{41} + 0.0298\cdot x_{108} - 0.014\cdot x_{111} - 0.735\cdot e^{-e^{(x_{19})}} + 0.753\cdot e^{{(-e^{(-x_{4})^{1/2}})}^{1/2}} - 3.52\cdot10^{-6}\cdot {x_{111}}^{3}\cdot {x_{4} + x_{5} + x_{21}}^{3} - 0.175\cdot e^{{(-e^{(-x_{21})})}^{1/2}} - 0.0498$
\\

$Y_{3}^{9} = 0.106\cdot x_{4} + 0.125\cdot x_{5} + 0.689\cdot x_{6} + 0.0261\cdot x_{11} + 0.45\cdot x_{19} + 0.0515\cdot x_{41} + 0.00826\cdot x_{58} - 0.744\cdot e^{(-e^{(x_{19})})} + 2.08\cdot10^{-7} e^{(2\cdot x_{97})}\cdot |x_{111}|^{2} + 0.00826\cdot x_{5}\cdot x_{6} - 0.00826\cdot x_{5}\cdot x_{28} - 0.00511\cdot x_{4}\cdot x_{74} + 0.00826\cdot x_{38}\cdot x_{74} + 0.00511\cdot {x_{4}}^{2} e^{(-x_{4})} - 0.176\cdot e^{({-x_{4})}^{1/2}} + 0.467$
\\

$Y_{3}^{10} = 0.0825\cdot x_{4} + 0.134\cdot x_{5} + 0.688\cdot x_{6} + 0.0174\cdot x_{12} + 0.47\cdot x_{19} + 0.0304\cdot x_{26} + 0.0409\cdot x_{41} - 0.523\cdot e^{(-e^{(-e^{(-x_{4})})})} - 0.703\cdot e^{(-e^{(x_{19})})} + 2.93\cdot{10}^{-4}\cdot {x_{72}}^{3} e^{(-x_{1})} + 0.635$

\newpage
In the figures below we show the model predictions against the actual values of the EGT per model on the test set:

\begin{figure}[h]
\centering
\subfloat[Model 1]{\includegraphics[width = 3in]{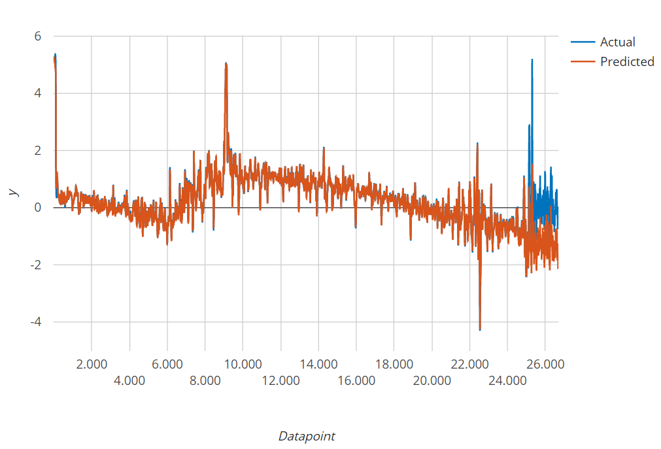}} 
\subfloat[Model 2]{\includegraphics[width = 3in]{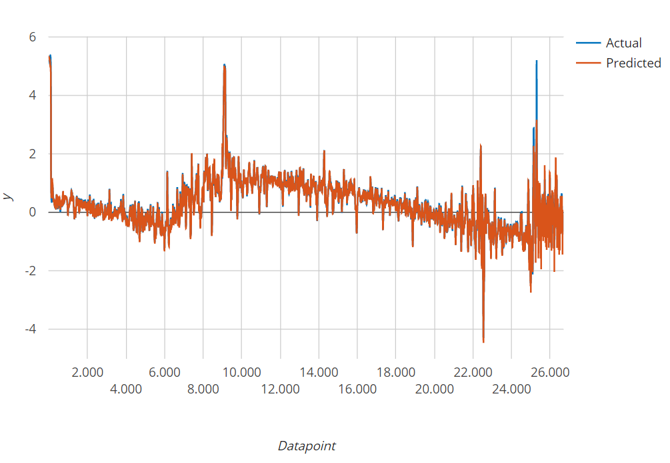}}\\ 
\subfloat[Model 3]{\includegraphics[width = 3in]{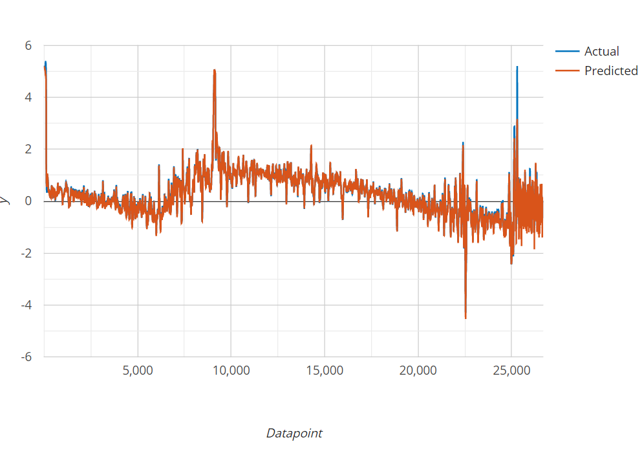}}
\subfloat[Model 4]{\includegraphics[width = 3in]{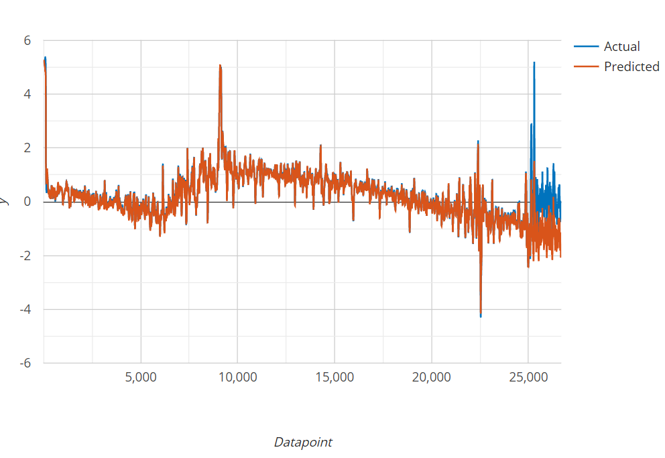}}\\

\caption{Scaled EGT predictions (red) ($y$-axis) on the validation set 
    vs.~observed (blue) values (Experiment 3). $x$-axis shows the data sample index in consecutive order according to their sampling over time.}
\label{ModelsExp3.1-4}
\end{figure}

\begin{figure}[!h]
\centering
\subfloat[Model 5]{\includegraphics[width = 3in]{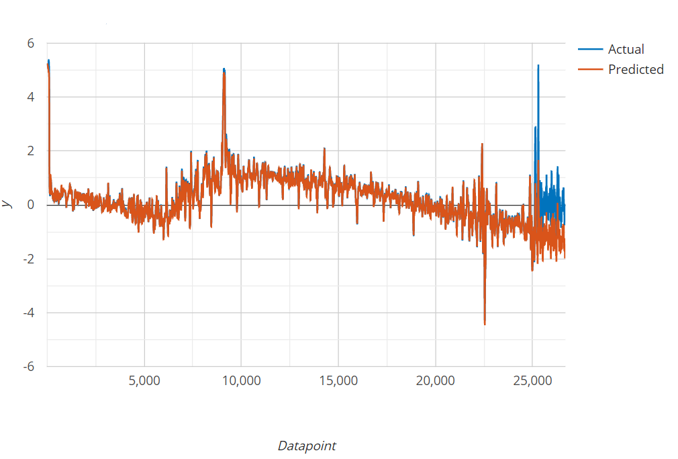}}
\subfloat[Model 6]{\includegraphics[width = 3in]{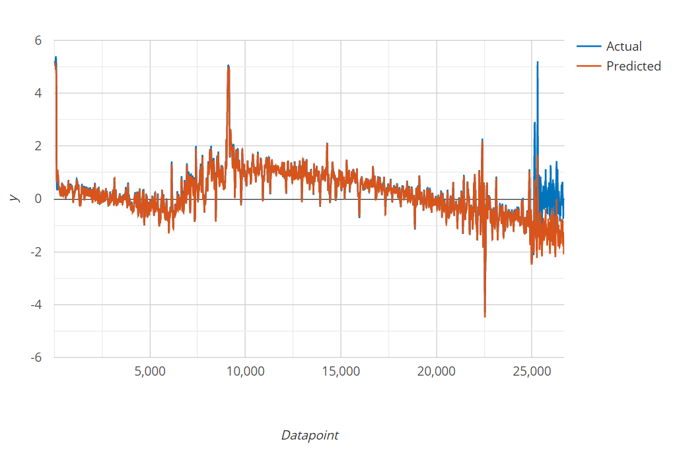}}\\
\subfloat[Model 7]{\includegraphics[width = 3in]{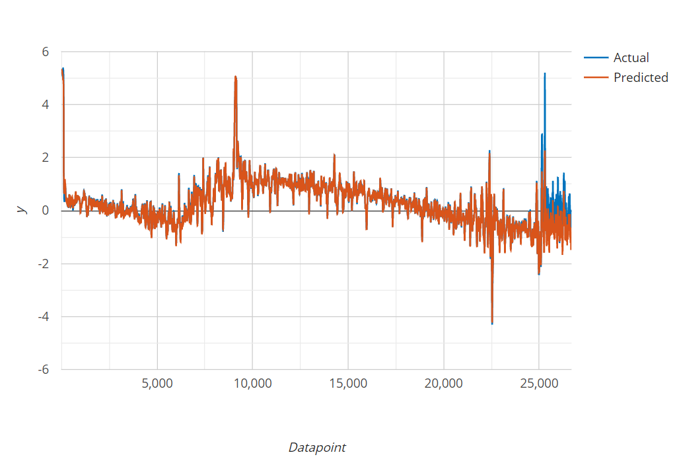}}
\subfloat[Model 8]{\includegraphics[width = 3in]{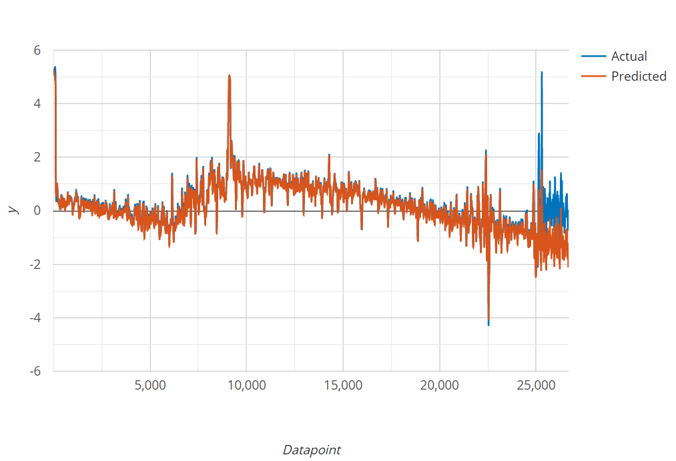}}\\
\subfloat[Model 9]{\includegraphics[width = 3in]{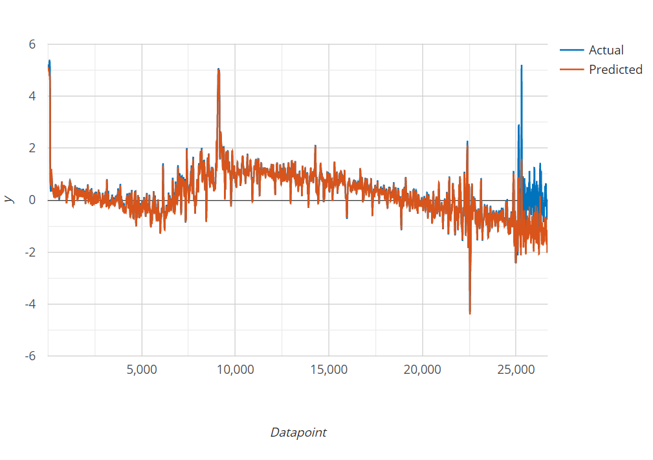}}
\subfloat[Model 10]{\includegraphics[width = 3in]{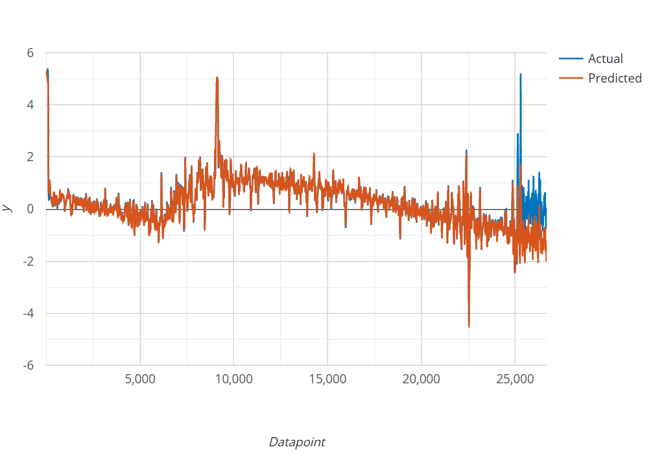}}
\caption{Scaled EGT predictions (red) ($y$-axis) on the validation set 
    vs.~observed (blue) values (Experiment 3). $x$-axis shows the data sample index in consecutive order according to their sampling over time.}
\label{ModelsExp3.5-10}
\end{figure}

\newpage
\section{Conclusion}\label{sec:conclusion}

This document is supplemental material to the main paper ``An Application of Symbolic Regression on Exhaust Gas Temperature of Turbofan Engines: A Case Study'', by Kefalas, de Santiago Rojo Jr., Apostolidis, van den Herik, van Stein and B\"ack. 
It holds in detail all the results from the experiments performed, together with plots and explanations, as well as all the generated formulas.
From the results we can see that Symbolic Regression (SR) is able to model accurately the EGT in all three experimental scenarios.
We should note that the authors confirmed the engineering validity of the symbolic expression model of the main paper, while the work presented in the Supplemental Material was meant to be presented as an example of the process and the possible factors associated with the prediction of the EGT. 
Future work can address the validity of some of these additional expressions, especially if interesting correlations between some engine parameters and EGT are identified.

\section*{Funding Sources}
This work is part of the research programme Smart Industry SI2016 with project name CIMPLO and project number 15465, which is partly financed by the Netherlands Organisation for Scientific Research (NWO).

\bibliography{sup_mybibliography}